\documentclass{article}

\usepackage{microtype}
\usepackage{graphicx}
\usepackage{booktabs} 

\usepackage{hyperref} 

\usepackage[accepted]{icml2025}

\usepackage{amsmath}
\usepackage{amssymb}
\usepackage{mathtools}
\usepackage{amsthm}

\usepackage[capitalize,noabbrev]{cleveref}

\theoremstyle{plain}
\newtheorem{theorem}{Theorem}[section]

\theoremstyle{definition}
\newtheorem{definition}[theorem]{Definition}
\newtheorem{assumption}[theorem]{Assumption}
\newtheorem{example}[theorem]{Example}
\theoremstyle{remark}

\usepackage{nicefrac}       
\usepackage{physics}        
\newcommand{\iter}[2]{#1^{(#2)}}
\usepackage{xcolor}%
\usepackage{pdflscape}
\usepackage{multirow}
\usepackage{subcaption}
\usepackage{tabularx}

\usepackage[normalem]{ulem} 
\usepackage{blindtext}%
\usepackage[textsize=scriptsize]{todonotes}%
\presetkeys{todonotes}{fancyline}{}
\newcommand{\fd}[1]{{[\color{green!50!black} \textbf{FD:} {#1}]}}%
\usepackage{url}%

\usepackage{enumitem} 

\usepackage{amsmath}
\usepackage{amssymb}
\usepackage{bm}
\usepackage{mathtools}

\def\vzero{{\bm{0}}}
\def\vone{{\bm{1}}}
\def\valpha{{\bm{\alpha}}}

\def\vtheta{{\bm{\theta}}}

\def\vdelta{{\bm{\delta}}}

\def\vb{{\bm{b}}}
\def\vc{{\bm{c}}}

\def\vpsi{{\bm{\psi}}}

\def\vr{{\bm{r}}}

\def\vv{{\bm{v}}}

\def\vw{{\bm{w}}}
\def\vx{{\bm{x}}}
\def\vy{{\bm{y}}}

\def\vz{{\bm{z}}}

\def\evtheta{{\theta}}

\def\mA{{\bm{A}}}
\def\mB{{\bm{B}}}

\def\mG{{\bm{G}}}

\def\mH{{\bm{H}}}
\def\mI{{\bm{I}}}

\def\mL{{\bm{L}}}

\def\mR{{\bm{R}}}

\def\mW{{\bm{W}}}
\def\mX{{\bm{X}}}

\def\mLambda{{\bm{\Lambda}}}

\DeclareMathAlphabet{\mathsfit}{\encodingdefault}{\sfdefault}{m}{sl}
\SetMathAlphabet{\mathsfit}{bold}{\encodingdefault}{\sfdefault}{bx}{n}

\def\gA{{\mathcal{A}}}

\def\gE{{\mathcal{E}}}

\def\gG{{\mathcal{G}}}
\def\gH{{\mathcal{H}}}

\def\gM{{\mathcal{M}}}

\def\gV{{\mathcal{V}}}

\def\sD{{\mathbb{D}}}

\def\sR{{\mathbb{R}}}
\def\sS{{\mathbb{S}}}

\DeclareSymbolFont{bbold}{U}{bbold}{m}{n}
\DeclareSymbolFontAlphabet{\mathbbold}{bbold}

\newcommand{\E}{\mathbb{E}}

\DeclareMathOperator{\GL}{GL}

\let\grad\relax 
\DeclareMathOperator{\grad}{grad}

\DeclareMathOperator{\orbit}{orbit}

\usepackage{booktabs}
\usepackage{hyperref}
\usepackage{url}
\usepackage{algpseudocode}
\usepackage{xparse} 
\usepackage{float}

\newcommand{\qDecorate}[1]{#1}
\newcommand{\tDecorate}[1]{\bar{#1}}

\NewDocumentCommand{\qTanVec}{ O{\xi} O{x}}{\qDecorate{#1}_{\qDecorate{#2}}}
\NewDocumentCommand{\tTanVec}{ O{\xi} O{x}}{\tDecorate{#1}_{\tDecorate{#2}}}

\NewDocumentCommand{\tHorVec}{ O{\xi} O{x}}{\tTanVec[#1][#2]^{\gH}}
\NewDocumentCommand{\tVertVec}{ O{\xi} O{x}}{\tTanVec[#1][#2]^{\gV}}

\begin{document}

\newcommand{\papertitle}{%
Hide \& Seek: Transformer Symmetries Obscure Sharpness \& Riemannian Geometry Finds It
}%

\newcommand{\sageevskip}[1]{} 

\icmltitlerunning{\papertitle}

\twocolumn[
\icmltitle{\papertitle}



\icmlsetsymbol{equal}{*}

\begin{icmlauthorlist}
\icmlauthor{Marvin F. da Silva}{dal,vec}
\icmlauthor{Felix Dangel}{vec}
\icmlauthor{Sageev Oore}{dal,vec}
\end{icmlauthorlist}

\icmlaffiliation{dal}{Faculty of Computer Science, Dalhousie University, Halifax, Canada}
\icmlaffiliation{vec}{Vector Insitute for Artificial Intelligence, Toronto, Canada}

\icmlcorrespondingauthor{Marvin F.da Silva}{marvinf.silva@dal.ca}

\icmlkeywords{Machine Learning, ICML}

\vskip 0.3in
]


\printAffiliationsAndNotice{}  

\begin{abstract}
The concept of sharpness has been successfully applied to traditional architectures like MLPs and CNNs to predict their generalization. For transformers, however, recent work reported weak correlation between flatness and generalization. We argue that existing sharpness measures fail for transformers, because they have much richer symmetries in their attention mechanism that induce directions in parameter space along which the network or its loss remain identical. We posit that sharpness must account fully for these symmetries, and thus we redefine it on a quotient manifold that results from quotienting out the transformer symmetries, thereby removing their ambiguities. Leveraging tools from Riemannian geometry, we propose a fully general notion of sharpness, in terms of a geodesic ball on the symmetry-corrected quotient manifold. In practice, we need to resort to approximating the geodesics. Doing so up to first order yields existing adaptive sharpness measures, and we demonstrate that including higher-order terms is crucial to recover correlation with generalization. We present results on diagonal networks with synthetic data, and show that our geodesic sharpness reveals strong correlation for real-world transformers on both text and image classification tasks.
\end{abstract}

\section{Introduction}
Predicting generalization of neural nets (NNs)---the discrepancy between training and test set performance---remains an open challenge.
Generalization-predictive metrics are valuable though: they enable explicit regularization of training to enhance generalization \citep{foret2021sharpnessaware}, and provide broader theoretical insights into generalization itself.

There is a long history of hypotheses linking sharpness to generalization, but evidence has been conflicting~\citep{hochreiter1994simplifying,andriushchenko2023modern}.
Generalization has been speculated as correlating with flatness, but recent evidence has indicated that, in the case of transformers, it has little to no correlation whatsoever.
Measures of sharpness have varied widely, ranging from trace of the Hessian to worst-case loss within a local neighborhood, with adaptive and relative variations proposed to address specific challenges~\citep{kwon2021asam,petzka2021relative}.

We suspect that some of the confusion stems from the specificity of the problem these measures have attempted to address: the issue of parameter rescaling.
In contrast, we argue that rescaling~\citep{dinh2017sharp} is merely a special case of a broader, more fundamental obstacle to measuring sharpness accurately: the presence of full and continuous parameter symmetries.
Addressing this challenge is crucial to ensure that we are studying the right quantity when investigating the relationship between sharpness and generalization.

\begin{figure*}[!tb]
  \centering
  \begin{subfigure}[t]{0.35\linewidth}
    \centering
    \caption{Loss}
    \label{subfig:visualization-scalar-toy-loss}
    \includegraphics[trim=0.5cm 0cm 0cm 0cm,width=\linewidth]{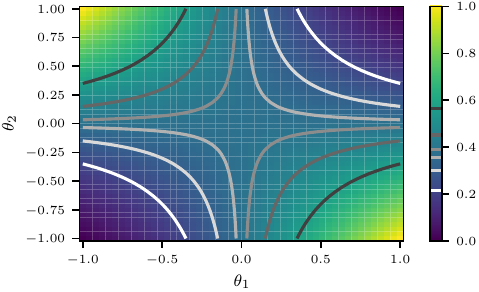}
  \end{subfigure}
  \begin{subfigure}[t]{0.3\linewidth}
    \centering
    \caption{Euclidean gradient norm}
    \label{subfig:visualization-scalar-toy-euclidean-gradient}
    \includegraphics[width=\linewidth]{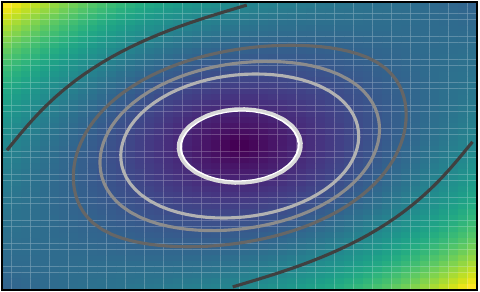}
  \end{subfigure}
  \begin{subfigure}[t]{0.3\linewidth}
    \centering
    \caption{Riemannian gradient norm}
    \label{subfig:visualization-scalar-toy-riemannian-gradient}
    \includegraphics[width=\linewidth]{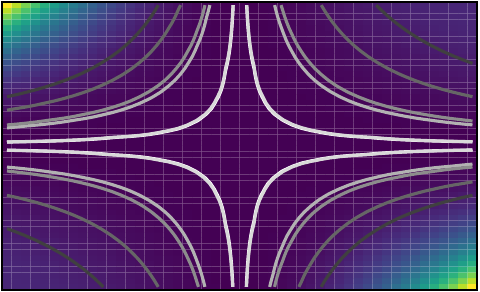}
  \end{subfigure}

  \caption{
    \textbf{Quantities from the Riemannian quotient manifold respect the loss landscape's symmetry; Euclidean quantities do not.}
    We illustrate this here for a synthetic least squares regression task with a two-layer NN, where $x \mapsto \theta_2 \theta_1 x$ with scalar parameters $\vtheta \in \sR^2$ and input $x \in \sR$ (i.e. each layer is a linear function).
    The NN is re-scale invariant, i.e.\,has $\GL(1)$ symmetry: For any $\alpha \in \sR \setminus \{ 0 \}$, the parameters $(\theta_1', \theta_2') = (\alpha^{-1}\theta_1, \alpha \theta_2)$ represent the same function.
    (\subref{subfig:visualization-scalar-toy-loss}) The loss function inherits this symmetry and has hyperbolic level sets.
    (\subref{subfig:visualization-scalar-toy-euclidean-gradient}) The Euclidean gradient norm does not share the loss function's geometry and changes throughout an orbit where the NN function remains constant.
    (\subref{subfig:visualization-scalar-toy-riemannian-gradient}) The Riemannian gradient norm follows the loss function's symmetry and remains constant throughout an orbit, i.e., it does not suffer from ambiguities for two points in parameter space that represent the same NN function.
  }
  \label{fig:visualization-scalar-toy}
\end{figure*}

Beyond discrete permutation symmetries, neural nets naturally exhibit continuous symmetries in their parameter space.
These symmetries are intrinsic, data-independent properties that emerge from standard architectural components.
For example: normalization layers~\citep{ioffe2015batch,ba2016layer,wu2018group} induce scale invariance on the pre-normalization weights~\citep{salimans2016weight}; homogeneous activation functions like $\mathrm{ReLU}$ introduce re-scaling symmetries between pre- and post-activation weights~\citep{dinh2017sharp}; some normalization layers and softmax impose translation symmetries in the preceding layer's biases~\citep{kunin2021symmetry}.
As a result, arguably almost any NN, along with its corresponding loss, exhibit symmetries and can therefore represent the \emph{same} function using \emph{different} parameter values (\Cref{subfig:visualization-scalar-toy-loss}).


Adaptive flatness \citep{kwon2021asam} accounts for some symmetries, both element- and filter-wise, but fails to capture the attention mechanism's \emph{full} symmetry, represented by $\GL(h)$ (re-scaling by invertible $h\times h$ matrices, where $h$ is the hidden dimension), as we will discuss later.
Aiming to break the cycle between discovery of a specific symmetry and techniques to deal with it, we ask:

\emph{Can we provide a one-size-fits-many recipe for developing symmetry-invariant quantities for a wider range of symmetries?}

Here, we positively answer this question by proposing a principled approach to eliminate ambiguities stemming from symmetry.
Essentially, this boils down to using the geometry that correctly captures symmetry-imposed parameter equivalences.
We apply concepts from Riemannian geometry to work on the Riemannian quotient manifold implied by a symmetry group~\citep[][\S9]{boumal2023introduction}.
We thus identify objects on the quotient manifold---like the Riemannian metric and gradient---and show how to translate them back to the Euclidean space.
Our contributions are the following:
\begin{enumerate}[
	label=(\alph*), 
	itemsep=0.25\parsep, 
	topsep=0.0\parsep, 
	]

\item We introduce the application of Riemannian geometry~\citep{boumal2023introduction} to the study of NN parameter space symmetries by using geometry from the quotient manifold induced by a symmetry as a general recipe to remove symmetry-induced ambiguities in parameter space.
  We do so by translating concepts like gradients from the quotient manifold back to the original space through \emph{horizontal lifts}.
  \label{item:first-contribution}
  \vspace{-0.1cm}
\item Based on \ref{item:first-contribution}, we propose and analyze \textit{geodesic sharpness}, a novel adaptive sharpness measure:
  By Taylor-expanding our refined geometry, we show that (i) symmetries introduce curvature into the parameter space, which (ii) results in previous adaptive sharpness measures when ignored.
  Geodesic sharpness differs from traditional sharpness measures in two key aspects: i) the norm of the perturbation parameter is redefined to reflect the underlying geometry; ii) perturbations follow geodesic paths in the quotient manifold rather than straight lines in the ambient space.
  \vspace{-0.1cm}
\item For diagonal nets, we analytically solve \emph{geodesic sharpness} and find a strong correlation with generalization.
  Then, we apply our approach to the unstudied and higher-dimensional $\GL(h)$ symmetry in the attention mechanism.
  On both large vision transformers and language models, we empirically find stronger correlation than any previously seen (that we are aware of) between our geodesic sharpness and generalization.
\end{enumerate}


\section{Related Work}
\paragraph{Symmetry vs.\,reparameterization:} \citet{kristiadi2023geometry} pointed out how to fix ambiguities stemming from
reparameterization, i.e.\,a change of variables to a \emph{new} parameter space: Invariance under reparameterization follows by correctly transforming the (often implicitly treated) Riemannian metric, into the new coordinates.

Our work focuses on invariance of the parameter space $\overline{\gM}$ under a symmetry group $\gG$ with action $\vpsi: \gG \times \overline{\gM} \to \overline{\gM}, \ (g, \vtheta) \mapsto \vpsi(g, \vtheta)$ that operates on a \emph{single} parameter space.

\paragraph{Symmetry teleportation:} Other ways to circumvent the ambiguity is to view it as a degree of freedom and develop adaptation heuristics to improve algorithms which are not symmetry-agnostic~\citep{zhao2022symmetry}. 

\paragraph{Geometric constraints \& NN dynamics:}

Previous studies analyze symmetries in parameter spaces by imposing geometric constraints on derivatives or identifying conserved quantities during training \citep{kunin2021symmetry}. Our approach differs by systematically removing symmetry-induced ambiguity through quotienting the parameter space by the symmetry group.

We generalize earlier post-hoc solutions for simpler symmetries (e.g., $\GL(1)$) to more complex, higher-dimensional symmetries such as $\GL(h)$, common in neural network attention mechanisms. Unlike \citet{kunin2021symmetry}, who consider geometry in augmented spaces for simpler symmetries, we directly use the quotient space geometry. Objects are then 'lifted' back into the original space, yielding symmetry-corrected gradients and Hessians. This method provides a principled framework capable of handling high-dimensional symmetries for a more effective dimensionality reduction.
\paragraph{Quotient manifolds in deep neural networks:}
\citet{rangamani2019scale} introduces a quotient manifold construction for re-scaling symmetries and then use the Riemannian spectral norm as a measure of worst-case flatness. This differs from our approach in several key ways:
\begin{enumerate}[
	label=(\alph*), 
	itemsep=0.0\parsep, 
	topsep=0.0\parsep, 
	]
    \item Our approach is more general and can accommodate both the $GL(h)$ symmetry of transformers, and the original re-scaling/scaling symmetry of convolutional/fully-connected networks, rendering it applicable to a wider range of modern architectures;
\item Our experimental setup is more challenging: we test on large-scale models (large transformers vs CNNs) and large-scale datasets (ImageNet vs CIFAR-10). Sharpness measures that account for re-scaling/scaling symmetries (e.g. adaptive sharpness) work quite well on CIFAR-10 and for CNNs and tends to break down on datasets like ImageNet and for transformers;
\item Conceptually, \citet{rangamani2019scale} defines worst-case sharpness on the usual norm-ball, appropriately generalized to the Riemannian setting. We propose instead that the ball should be the one traced out by geodesics, to better respect the underlying geometry. 
\item Performance-wise, our approach is more efficient because it does not use the Hessian and needs only to use considerably cheaper gradients (for a more in-depth comparison of the cost of Hessians vs gradients see, e.g. \citet{dagreou2024howtocompute}).
\end{enumerate}
\paragraph{Relative sharpness:} Another promising approach to sharpness was proposed by \citet{petzka2021relative}, where the generalization gap is shown to admit a decomposition into a representativeness term and a feature robustness term. Focusing on the feature robustness term they introduce relative sharpness, which is invariant to a layer and neuron-wise network re-scalings, and performs better than traditional sharpness measures (\citet{adilova2023famrelativeflatnessaware},\citet{walter2025uncannyvalley}).


\section{Preliminary Definitions, Notation \& Math}
\paragraph{Generalization measures:} We consider a neural net $f_\vtheta$ with parameters $\vtheta \in \sR^D$ that is trained on a data set $\sD_\text{train}$ using a loss function $\ell$ by minimizing  empirical risk
\begin{align*}
	L_{\sD_{\text{train}}}(\vw) \coloneqq \frac{1}{|\sD_{\text{train}}|}
	\textstyle
	\sum_{(\vx, \vy) \in \sD}
	\ell(f_\vtheta(\vx), \vy) \,.
\end{align*}
Our goal is to compute a quantity on the training data that is predictive of the network's generalization, i.e.\,performance on a held-out data set.

\vspace{-0.3cm}
\paragraph{Sharpness:} A popular way to predict generalization is via sharpness, i.e., how much the loss changes when perturbing the weights in a small neighbourhood.
E.g., one can consider the average ($S_{\text{avg}}$) or worst-case sharpness ($S_{\text{max}}$)

\begin{align*}
	\begin{split}
		\hspace{-0.2cm} S_{\text{avg}}
		&=
		\E_{\sS} \left[ L_{\sS}(\vtheta + \vdelta) - L_{\sS}(\vtheta) \right], \quad \vdelta \sim \mathcal{N}(\vzero, \rho^2\mI)\,,
		\\
		\hspace{-0.2cm} S_{\text{max}}
		&=
		\E_{\sS} \left[ \max_{\|\vdelta\|_2 \leq \rho} \left( L_{\sS}(\vtheta + \vdelta) - L_{\sS}(\vtheta) \right) \right]\,,
	\end{split}
\end{align*}

Near critical points, they closely relate to the Hessian $\mH$ (and thus parameter space curvature): $S_{\text{avg}} \sim \Tr(\mH)$, $S_{\text{max}} \sim \lambda_{\text{max}}(\mH)$.

\vspace{-0.3cm}
\paragraph{Adaptive sharpness:} Hessian-based sharpness measures can be made to assume arbitrary values by rescaling parameters, even though the NN function stays the same.
To fix this inconsistency, \citet{kwon2021asam} proposed adaptive sharpness (invariant under special symmetries), and \citet{andriushchenko2023modern} use adaptive notions of sharpness that are invariant to element-wise scaling,
\begin{align} \label{eq:max_def}
	\hspace{-0.2cm}
	S_{\text{max}}^{\rho}(\vw, \bm{c})
	=
	\E_{\sS} \left[ \max_{\|\vdelta \odot \vc^{-1}\|_2 \leq \rho} L_{\sS }(\vtheta + \vdelta) - L_{\sS}(\vtheta) \right],
\end{align}
with scaling vector $\bm{c}$ \citep[usually set to $|\vtheta|$,][]{kwon2021asam}.
\vspace{-0.3cm}
\paragraph{The problem:} Adaptive sharpness only considers a special symmetry.
But symmetries of transformers go beyond the invariance that adaptive sharpness captures.
Maybe unsurprisingly, \citet{andriushchenko2023modern} find inconsistent trends for adaptive sharpness in transformers versus other architectures.
We hypothesize this is related to adaptive sharpness not accounting for the full symmetry in transformers.
In this paper, we address this.
The central question is:
\textit{If adaptive sharpness is the fix for a special symmetry, can we provide a more general solution for the symmetries of transformers, to fix the above inconsistency?}

\subsection{Symmetries in Neural Networks}\label{subsec:symmetries-in-nns}
Here, we give a brief overview and make more concrete the notion of NN symmetries, focusing on those studied by \citet{kunin2021symmetry}.
Those symmetries lead to rather small effective dimensionality reduction as they are often of $\GL(1)$ or $\GL^+(1)$, but they can still impact the network behaviour considerably.
Let $\vtheta$ denote the parameters of a neural network, $\vone_{\gA}$ a binary mask, and $\vone_{\neg \gA}$ its complement such that their sum is a vector of ones, $\vone_{\gA} + \vone_{\neg \gA} = \vone$.
Let $\vtheta_{\gA} \coloneq \vtheta \odot \vone_{\gA}$ with $\odot$ the element-wise product.
Further, let $\gA_{1,2}$ be two disjoint subsets, $\gA_1 \cap \gA_2 = \emptyset$ with masks $\vone_{\gA_1}, \vone_{\gA_2}$.
Then we have the following common symmetries, characterized by their symmetry group $\gG$, such that for any $g \in \gG$ the parameter $\psi(g, \vtheta)$ represents the same function as $\vtheta$:
\begin{itemize}[nosep]
\item \textbf{Translation:}
  $\psi(\valpha, \vtheta) = \vone_{\gA} \odot \valpha + \vtheta$ with $\valpha \in \sR^h$
\item \textbf{Scaling:}
  $\psi(\alpha, \vtheta) = \alpha \vtheta_{\gA} + \vtheta_{\neg \gA}$ with $\alpha \in \sR_{>0}$
\item \textbf{Re-scaling:}
  $\psi(\alpha, \vtheta) = \alpha \vtheta_{\gA_1} +  \nicefrac{1}{\alpha} \vtheta_{\gA_2} + \vtheta_{\neg (\gA_1 \lor \gA_2)}$ with $\alpha \in \sR_{>0}$
\end{itemize}
Their associated groups are $\gG = \sR^h, \GL^{+}(1), \GL^{+}(1)$.
In practice, there may be multiple symmetries acting onto disjoint parameter sub-spaces.
Note that the re-scaling symmetry is essentially the symmetry that adaptive sharpness corrects for.

\subsection{Rescale Symmetry of Transformers}\label{subsec:symmetries-in-transformers}
Transformers exhibit a higher-dimensional symmetry than the previous examples; we formalize the treatment of this symmetry in the following canonical form.

\begin{definition}[Functional $\GL$-symmetric building block]\label{def:building-block-gl-symmetry}
  Consider a function $\overline{\overline{f}}(\mG, \mH)$ on $\sR^{m \times h} \times \sR^{n \times h}$ that consumes two matrices $\mG \in \sR^{n \times h}, \mH \in \sR^{m \times h}$ but only uses the product $\mG \mH^{\top}$, i.e.\,$\overline{\overline{f}}(\mG, \mH) = g(\mG\mH^{\top})$ for some $g$ over $\sR^{m \times n}$.
  $\overline{\overline{f}}$ is symmetric under the \emph{general linear group}
  \begin{subequations}
    \begin{align*}
      &\GL(h) \coloneqq \left\{ \mA \in \sR^{h \times h} \mid \mA\,\text{invertible} \right\},
               \shortintertext{with $\dim(\GL(h)) = h^2$ and action}
               &\psi(\mA, (\mG, \mH)) = (\mG \mA^{-1}, \mH \mA^{\top})\,.
               \label{eq:rescaleMatrices}
    \end{align*}
  \end{subequations}
  In other words, we can insert and then absorb the identity $\mA^{-1}\mA$ into $\mG, \mH$ to obtain equivalent parameters $\mG \mA^{-1}, \mH \mA^{\top}$ that represent the same function.
\end{definition}
\cref{ap:ex:shallow-linear-net} illustrates $\GL$ symmetry for a shallow linear net.
Indeed, many popular NN building blocks feature this form--most prominently the attention mechanism in transformers~\cite{vaswani2017attention}.
We give the attention symmetry in Example~\ref{ex:self-attention}, and we provide the symmetry for low-rank adapters~\citep{hu2022lora} in \cref{ap:ex:lora}.
These examples are NN building blocks that introduce $\GL$ symmetries into a loss function produced and can all be treated through the canonical form in \Cref{def:building-block-gl-symmetry}.
In contrast to the symmetries from \Cref{subsec:symmetries-in-nns}, they lead to more drastic dimensionality reduction.
Consider for example a single self-attention layer where $d = d_{\text{v}} = d_{\text{k}}$.
The number of trainable parameters is $4 d^2$ and the two $\GL(d)$ symmetries reduce the effective dimension to $4d^2 - 2 \dim(\GL(d)) = 2d^2$, i.e.\,they render \emph{half} the parameter space redundant.
We hypothesize that the range of objects like the Euclidean Hessian's trace~\citep{dinh2017sharp} in the presence of a low-dimensional symmetry may be amplified for such higher-dimensional symmetries.

\subsection{Mathematical Concepts for Riemannian Geometry}

We now outline required properties of manifolds for the full development of our approach.
We list essential concepts here, and provide definitions and a brief review \cref{ap:mathdefs}.
For further information, see for instance\,\citet{lee2003introduction}.

\begin{figure*}[!h]
  \centering
  \begin{subfigure}[]{0.495\linewidth}
    \centering
    \includegraphics[width=\linewidth]{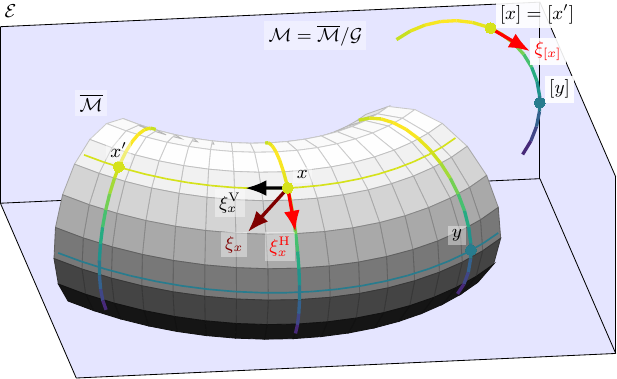}
  \end{subfigure}
  \hfill
  \begin{subfigure}[t]{0.45\linewidth}
    \centering
    \begin{small}
    \begin{tabularx}{\linewidth}{cX}
      $\gE$ & Linear embedding space
      \\
      $\overline{\gM}$ & Total space
      \\
      $\gM$ & Quotient space
      \\
      $\gG$ & Symmetry group
      \\
      $\bar{x}, \bar{y}$ & Points on the total space
      \\
      $x, y$ & Points on the quotient space
      \\
      $\bar{\xi}_{\bar{x}}$ & Tangent vector in the tangent space at point $\bar{x}$, $\mathrm{T}_{\bar{x}} \overline{\gM}$
      \\
      $\xi_{x}$ & Tangent vector in the tangent space at point $x$, $\mathrm{T}_{x} \gM$
      \\
      $\bar{\xi}_{\bar{x}}^{\mathrm{V}}$ & Vertical component of $\bar{\xi}_{\bar{x}}$ in the vertical space $\mathrm{V}_{\bar{x}} \overline{\gM}$
      \\
      $\bar{\xi}_{\bar{x}}^{\mathrm{H}}$ & Horizontal component of $\bar{\xi}_{\bar{x}}$ in the horizontal space
      $\mathrm{H}_{\bar{x}} \overline{\gM} \simeq \mathrm{T}_{x}\gM$, horizontal lift of $\xi_{x}$
      \end{tabularx}
    \end{small}
  \end{subfigure}
  \caption{
    \textbf{Illustrative sketch relating total and quotient space and their tangent spaces.}
    A tangent vector at a point in total space, $\tTanVec \in T_{\bar{x}} \overline{\gM}$ can be decomposed into a horizontal component $\tHorVec$ and a vertical component $\tVertVec$.
    The vertical component points along the direction where the quotient space $\qDecorate{x} = [\tDecorate{x}]$ remains unaffected.
    The horizontal component points along the direction that changes the equivalence class.
    We can use $\tHorVec$ as a representation of the tangent vector $\qTanVec \in T_x \gM$ on the quotient space.
    The component $\tHorVec$ represents the \emph{horizontal lift} of $\qTanVec$.}
  \label{fig:sketch}
\end{figure*}

\paragraph{Ambient embedding space} We assume our parameter manifold to be embedded in a linear Euclidean space $\gE \simeq \sR^d$ with $d$ the number of parameters.
We can think of $\gE$ as the \emph{computation space}.
For instance, for a loss function $\overline{\overline{\ell}}: \gE \to \sR, \vtheta \mapsto \overline{\overline{\ell}}(\vtheta)$ , we can use ML libraries to evaluate its value, as well as its Euclidean gradient \vspace{-0.1cm}
\begin{equation*}
    \grad_{\vtheta} \overline{\overline{\ell}}
    =
      \left(
      \frac{\partial \overline{\overline{\ell}}(\vtheta)}{\partial \evtheta_i}
      \right)_{i=1,\dots, d} \in \sR^d .
\end{equation*}
Because the geometry of $\gE$ is flat, i.e.\,uses the standard metric $\langle \vtheta_1, \vtheta_2 \rangle \coloneqq \vtheta_1^{\top} \vtheta_2$, this object consists of partial derivatives.
However, the Riemannian generalization will add correction terms. In what follows we consider only the restriction of objects like $\overline{\overline{\ell}}$ to the parameter space.

\begin{definition}
	We take $\overline{\gM}$ to be the manifold of network parameters, and consider it a sub-manifold embedded into $\gE$, the computational space of matrices on which all our numerical calculations are done.
  We call $\overline{\gM}$ the \emph{total space}.
  On the total space we have a loss function $\overline{\ell}: \overline{\gM} \rightarrow \sR$.
\end{definition}
Our goal is to calculate derivatives/geometric quantities after removing the NN's symmetries.
The symmetry relation induces natural equivalence classes, which we write $[x]$, and explain in \cref{ap:mathdefs:equiv}.
We let $\gM = \overline{\gM}/\sim \quad$ represent the \textit{quotient} of the original parameter space manifold by the equivalence relation associated with the symmetry (\cref{ap:mathdefs:quotient}).
We also require \textit{tangent vectors}; these are straightforward on the total space $\overline{\gM}$, but the tangent space of the quotient manifold, $\gM$, requires more machinery: \textit{vertical} and \textit{horizontal spaces}, and corresponding \textit{lift}s.
These concepts are all defined in \cref{ap:mathdefs:tangents}.

Once we endow $\overline{\gM}$ with a smooth inner product over its tangent vectors, we obtain a {\it Riemannian manifold} (defined in \cref{ap:mathdefs:riemannian}).
This construction lets us analyze differential objects that live on quotient manifolds, on the ambient space in a natural way.
Furthermore, this allows us to define the horizontal space as the orthogonal complement of the vertical space (\cref{ap:mathdefs:riemannian}), and to define a {\it Riemannian gradient} (\cref{ap:mathdefs:rgrad}).
Most properties from the Euclidean case are still true for the Riemannian gradient, but of particular interest to us is the fact that the direction of $\text{grad} f(x)$ is still the steepest-ascent direction of $f$ at a point $x$.

We additionally make use of {\it geodesic curves}.
Intuitively, geodesic curves can either be seen as curves of minimal distance between two points on a manifold $\overline{\gM}$, or equivalently, as curves through a given point with some initial velocity, and whose acceleration is zero--- a generalization of Euclidean straight lines.
See \cref{ap:mathdefs:geodesics} for details.

\vspace{-0.4cm}

\section{Geodesic Sharpness}\label{sec:geodesic_sharpness}
\newcommand{\uv}{\bm{u}}
\newcommand{\yv}{\bm{y}}
\newcommand{\xv}{\bm{x}}
\newcommand{\Xm}{\bm{X}}
\newcommand{\bv}{\bm{\beta}}
\newcommand{\Bv}{\bm{B}}
\newcommand{\Dv}{\bm{D}_{\bv_0,\bv_*}}
\vspace{-0.1cm}
We posit that adaptive sharpness measures should take into account the geometry of the quotient parameter manifold, that arises after removing symmetries from the parameter space.
We base our sharpness measure on the notion of a \emph{geodesic ball}: the set of points that can be reached by geodesics starting at a point $p$ and whose initial velocity has a norm smaller than $\rho$.

In $\mathbb{R}$this is just the usual definition of a ball, since the geodesics are straight lines.
If $\bar{\xi} \in H_w$
is a horizontal vector, and  $\bar{\gamma}(t)$ is a geodesic starting at $w$ and with initial velocity $\bar{\xi}$:
\vspace{-0.2cm}
\begin{align}\label{eq:max-sharpness}
\begin{split}
	S_{\text{max}}^{\rho}(\vw)
  &=
    \E_{\sS} \left[ \max_{\| \bar{\xi}\|_{\bar{\gamma}(0)} \leq \rho} L_{\sS}(\bar{\gamma}_{\bar{\xi}}(1)) - L_{\sS}(\bar{\gamma}_{\bar{\xi}}(0)) \right]
 \end{split}.
\end{align}
If the initial velocity, $\bar{\xi}$, is a horizontal vector, then the velocity of the geodesic, $\dot{\bar{\gamma}}_{\bar{\xi}}$, will stay horizontal. The choice of $t=1$ in $\bar{\gamma}_{\bar{\xi}}(1)$ is not as arbitrary as it seems~\citep{docarmo2016differential}: for a positive $a$,~$\bar{\gamma}_{\bar{\xi}}(a t)= \bar{\gamma}_{a \bar{\xi}}(t)$.

When we do not have an analytical solution for the geodesic, we can use the approximation:
\begin{align}\label{eq:geodesic_approximation}
    \bar{\gamma}_{\bar{\xi}}^i(t) = \bar{\gamma}_{\bar{\xi}}^i(0)+\bar{\xi}^i t -\frac{1}{2} \Gamma^i_{kl}\bar{\xi}^k\bar{\xi}^l t^2+\mathcal{O}(\bar{\xi}^3),
\end{align}
where $\bar{\xi} = (\bar{\xi}^i)$ is the initial (horizontal) velocity, and
$\Gamma^i_{kl}$ are the Christoffel symbols.

We show geodesic sharpness reduces to adaptive sharpness measures in Appendix~\ref{ap:ignoring_corrections}, under appropriate metric choices.

\vspace{-0.3cm}
\section{Geodesic Sharpness in Practice}
 In this section, we apply geodesic sharpness to concrete examples.
A fully worked out scalar toy model is provided in Appendix~\ref{ap:scalar_toy}.
Following previous works by \citet{dziugaite}, \citet{kwon2021asam}, \citet{andriushchenko2023modern}, we use the Kendall rank correlation coefficient~\citep{kendaltau} to assess the correlation between generalization and sharpness in the empirical validations of our approach:
\begin{align*}
  \tau(t, s) = \frac{2}{M(M-1)} \sum_{i<j} \text{sign}(t_i - t_j) \, \text{sign}(s_i - s_j),
\end{align*}
where $t$ and $s$ are the vectors between which we are trying to measure correlation. 

Although the criterion of symmetry compatibility greatly restricts the class of suitable metrics, these are not necessarily unique.
As long as this symmetry compatibility criterion is satisfied, we view the choice of metric as akin to a choice of preconditioning.
We will present results on two symmetry-compatible metrics that are simple, yet non-trivial, symmetry-compatible and often used in the related literature on Riemannian optimization on fixed-rank matrix spaces~\citep{luo2023quotient}.

\vspace{-0.3cm}
\subsection{Diagonal Networks} \label{subsec:diagonal_networks} \vspace{-0.1cm} We start by studying {\it diagonal linear nets}, one of the simplest non-trivial neural networks (\citet{pesme2021implicitbiassgddiagonal}, \citet{woodworth2020kernelrichregimesoverparametrized}).
These have two parameters, $\uv, \vv$, and predict a label, $\vy$, given an input, $\xv$, via $\yv = \xv^\top (\uv \odot \vv)$.
We consider a linear regression problem with labels ${\yv} \in \mathbb{R}^n$ and a data matrix ${\mX}\in\mathbb{R}^{n\times d}$.
We take as our loss $L({\uv},{\vv}) = \| {\mX} ({\uv} \odot {\vv}) - \yv\|^2_2$.
Our parameter manifold $\gM$ will be $\mathbb{R}^d \times \mathbb{R}^d$.

The nets are symmetric under element-wise rescaling: $(\uv,\vv) \mapsto (\alpha \uv,\alpha^{-1}\vv)$, leaves $\bv = \uv \odot \vv$, and hence the loss, invariant.
\vspace{-0.3cm}
\paragraph{Metric:} At a point $(\uv, \vv) \in \gM$, for two tangent vectors $\eta = (\eta_{\uv}, \eta_{\vv})$, $\nu = (\nu_{\uv}, \nu_{\vv}) \in T_{(\uv, \vv)} \gM$, we have the following two symmetry-compatible metrics

\begin{align}
&\langle\eta, \nu \rangle^{\text{inv}}
= \textstyle\sum_{i=1}^d \frac{\eta_{\uv}^i \nu_{\uv}^i}{(\uv^i)^2} +\frac{\eta_{\vv}^i \nu_{\vv}^i}{(\vv^i)^2},\label{eq:ginv_diagonal}
\\
&\langle\eta, \nu \rangle^{\text{mix}}
= \textstyle\sum_{i=1}^d \eta_{\uv}^i \nu_{\uv}^i (\vv^i)^2 +\eta_{\vv}^i \nu_{\vv}^i(\uv^i)^2. \label{eq:gmix_diagonal}
\end{align}
\vspace{-0.6cm}
\paragraph{Horizontal space:} The horizontal space is identical for both metrics

$H_{(\uv, \vv)} = \{\left(\eta_{\uv}, \eta_{\vv} \right) \in T_{(\uv, \vv)} \gM ~~|~~\frac{\eta_{\uv}^i}{\uv^i}=\frac{\eta_{\vv}^i}{\vv^i}~~\forall i \in \{1,\ldots, d\}\}$
 \vspace{-0.4cm}

 \paragraph{Geodesics:} We define $\Bv^i= \nicefrac{\eta_{\uv}^i}{\uv^i}=\nicefrac{\eta_{\vv}^i}{\vv^i} ~~\forall i \in \{1,\ldots, d\}$, so that
\begin{align*}
&\bm{\gamma}_{\text{inv}}(t)^i = \left(\uv_0^i \exp(\Bv_i t), \vv_0^i \exp(\Bv_i t) \right),
\\
&\bm{\gamma}_{\text{mix}}(t)^i = \left(\uv_0^i \sqrt{1+2\Bv_i t}, \vv_0^i \sqrt{1+2\Bv_i t} \right),
\end{align*}
where $\uv_0^i$ and $\vv_0^i$ are the initial positions of our parameters, that is, the parameters that the network actually learned.

\paragraph{Geodesic sharpness:}
Assume that $\Xm^\top\Xm = \mI_d$, and denote $\bv_0 = \uv_0 \odot \vv_0$.
The minimum norm least squares predictor is $\bv_* \coloneqq (\mX^{\top} \mX)^{-1} \mX^{\top} \vy =  \Xm^\top \vy$.

Using \eqref{eq:max-sharpness} (and leaving details to Appendix~\ref{ap:diagonal}), we get

\begin{align}
 	S_{\text{max; inv}}^{\rho}(\uv,\vv)
  &=
    4\rho \|\bv_0 \odot(\bv_0-\bv_{*})\|_2,
    \\
  S_{\text{max; inv}}^{\rho}(\uv,\vv)
  &=
    4 \rho^2 \max\left[(\beta_0^i)^2\right], \label{eq:adaptive_sharpness_diagonal}
\end{align}
depending on the value of $\rho$ and the difference between our learned predictor and the optimal min-norm one. Eq.~\ref{eq:adaptive_sharpness_diagonal} is the square of what one would obtain for adaptive sharpness if very carefully chosen hyperparameters were used (by contrast, this result naturally appears using our geodesic approach). For the second metric, we have
\begin{align*}
    S_{\text{max; mix}}^{\rho}(\uv,\vv) = \rho \|\bv_0-\bv_{*}\|_2  \, \, \, ,
\end{align*}
which is just the norm of the difference between our learned predictor and the optimal one.

\vspace{-0.3cm}
\subsection{Attention Layers}
\vspace{-0.1cm}

We take as our computation space
$\gE \coloneqq \sR^{n \times h} \times \sR^{m \times h} \simeq \sR^{(n + m) h}$.
In what follows we restrict our weight matrices to have full column rank.
\vspace{-0.1cm}
\begin{assumption}[]
  The rank of $\mG, \mH$ corresponds to their number of columns, $\rank(\mG) = \rank(\mH) = h$.
\end{assumption}
\vspace{-0.2cm} This implies $h \le n,m$, which is usually satisfied in (multi-head) attention layers (\Cref{ex:self-attention}) for the default choices of $d_{\text{v}}, d_{\text{k}}$.

While the weights of multi-head attention layers tend to have high column rank~\citep{yu2023lowrank}, they are not guaranteed to be full column rank.
To account for this we introduce a small relaxation parameter, $\epsilon$, s.t.
$\mG^\top \mG \rightarrow \mG^\top \mG+\epsilon \mI_h$.
Empirically, we observe that as long as $\epsilon$ is sufficiently small, it does not affect our results (\cref{ap:relaxation}).
In the following, we therefore restrict both $\mG, \mH$ to the set of fixed-rank matrices,
$\overline{\gM} \leftarrow \sR^{n \times h}_h \times \sR^{m \times h}_h$
where $\sR^{n \times h}_k \coloneqq \left\{\mB \in \sR^{n \times h} \mid \rank(\mB)=k \right\}$.

We can represent a point $\bar{x} \in \overline{\gM}$ by a matrix tuple $(\mG, \mH) \in \sR^{n \times h}_h \times \sR^{m \times h}_h$. Its tangent space $\mathrm{T}_{\bar{x}}\overline{\gM}$ is
\vspace{-0.3cm}
\begin{align*}
  \mathrm{T}_x\overline{\gM}
  =
  \left\{ \eta \in \sR^{n \times h} \times \sR^{m \times h} \right\},
\end{align*}
and a tangent vector $\eta \in \mathrm{T}_{\bar{x}}\overline{\gM}$ is represented by a matrix tuple $(\eta_{\mG}, \eta_{\mH}) \in \sR^{n \times h} \times \sR^{m \times h}$.
\vspace{-0.3cm}
\paragraph{Metric:} We endow $\overline{\gM}$ with the following two metrics $\langle  \cdot, \cdot \rangle_{\bar{x}}^{\text{inv,mix}}: \mathrm{T}_{\bar{x}} \overline{\gM} \times \mathrm{T}_{\bar{x}} \overline{\gM} \to \sR$ (proof that these define valid metrics in~\cref{ap:metric_proof}):
\vspace{-0.15cm}
\begin{small}%
  \begin{align}\label{eq:attention_metric}%
    \langle \bar{\eta}, \bar{\zeta} \rangle_{\bar{x}}^{\text{inv}} =
    \Tr
    \left(
    ( \mG^{\top}\mG )^{-1}
    \bar{\eta}_{\mG}^{\top} \bar{\zeta}_{\mG}
    +
    ( \mH^{\top}\mH )^{-1}
    \bar{\eta}_{\mH}^{\top} \bar{\zeta}_{\mH}
    \right)\,,
\end{align}
\end{small}%
\vspace{-0.8cm}
\begin{small}
  \begin{align}\label{eq:attention_metric_mix}%
    \langle \bar{\eta}, \bar{\zeta} \rangle_{\bar{x}}^{\text{mix}} =
    \Tr
    \left(
    ( \mH^{\top}\mH )
    \bar{\eta}_{\mG}^{\top} \bar{\zeta}_{\mG}
    +
    ( \mG^{\top}\mG )
    \bar{\eta}_{\mH}^{\top} \bar{\zeta}_{\mH}
    \right)\, \, \, .
  \end{align}%
\end{small}%
These are different from the Euclidean metric that simply flattens and concatenates the matrix tuples into vectors and takes their dot product, $\langle \eta, \zeta \rangle = \Tr \left( \eta_{\mG}^{\top} \zeta_{\mG} + \eta_{\mH}^{\top} \zeta_{\mH} \right)$. Importantly, these metrics are invariant under symmetries of the attention mechanism, and thus define valid metrics on the quotient manifold \citep{absil2008optimization}.
\paragraph{Horizontal space:} For $\langle  \cdot, \cdot \rangle_{\bar{x}}^{\text{inv, mix}}$ and $\bar{\xi}_{\mG,\mH} \in \mathbb{R}^{m \times r}$ we have

\begin{align*}
    &\mathcal{H}_{\bar{x}}^{\text{inv}}\overline{\gM} &&= \{ (\bar{\xi}_{\mG}, \bar{\xi}_{\mH}) \mid \bar{\xi}_{\mG}^\top \mG \mH^{T} \mH = \mG^{T} \mG  \mH^\top\xi_{\mH}^{T}\},
\\
    &\mathcal{H}_{\bar{x}}^{\text{mix}}\overline{\gM} &&= \{ (\bar{\xi}_{\mG}, \bar{\xi}_{\mH}) \mid \mG^{T} \bar{\xi}_{\mG} \mH^{T} \mH = \mG^{T} \mG \xi_{\mH}^{T} \mH \}.
\end{align*}
\vspace{-0.85cm}
\paragraph{Projection from total space onto horizontal space:} Given $ \xi
    \in  \mathrm{T}_x \overline{\gM}$ in the total tangent space
\begin{align*}
    \mathcal{H}_{\bar{x}}^{\text{inv, mix}} \overline{\gM} =
  \left\{(
    \bar{\xi}_{\mG} + \mG \mLambda^{\text{inv, mix}},
    \bar{\xi}_{\mH} -\mH (\mLambda^{\text{inv, mix}})^\top
    )
  \right\}.
\end{align*}
\vspace{-0.05cm}
$\bm{\Lambda}^{\text{inv}}$ is the solution of the Sylvester equation $\mA \mLambda + \mLambda \mA^{\top}  = \mB$, with $\mA = \mG^\top \mG \mH^\top \mH$,
$\mB = \mG^\top \mG \mH^\top \bar{\xi}_{\mH} - \bar{\xi}_{\mG}^{\top} \mG \mH^\top \mH$.

$\bm{\Lambda}^{\text{mix}}$, by contrast, has an explicit solution: $\bm{\Lambda}^{\text{mix}} = \frac{1}{2} \left(\bar{\xi}_{\mH}^\top \mH (\mH^\top \mH)^{-1} - (\mG^\top \mG)^{-1} \mG^\top \bar{\xi}_{\mG} \right)$
\vspace{-0.2cm}
\paragraph{Geodesics:} As far as we are aware there is no analytical solution for the geodesics of either metric (\ref{eq:attention_metric}) or (\ref{eq:attention_metric_mix}), so we use the approximation given by Eq.~\ref{eq:geodesic_approximation}.

For horizontal tangent vectors $(\bar{\xi}_G, \bar{\xi}_H)$
\begin{align}\label{eq:attention_approximation}
\begin{split}
    (\Gamma^i_{kl})^{\text{inv}} \bar{\xi}^k_\mG \bar{\xi}^l_\mG = &-\bar{\xi}_\mG(\mG^\top \mG)^{-1} \left[\bar{\xi}_\mG^\top \mG+\mG^\top \bar{\xi}_\mG\right]
    \\
    &+\mG(\mG^\top \mG)^{-1}\bar{\xi}_\mG^\top\bar{\xi}_\mG
    \end{split},
\end{align}
For $\langle  \cdot, \cdot \rangle_{\bar{x}}^{\text{mix}}$, the geodesic equations are coupled and the $\mG$ components are (the $\mH$ components are similar)
\begin{align}\label{eq:attention_approximation_mix}
\begin{split}
    \hspace{-0.2cm}\left[(\Gamma^i_{kl})^{\text{mix}} \bar{\xi}^k \bar{\xi}^l\right]_{\mG}  = &\bar{\xi}_{\mG}\left[\bar{\xi}_{\mH}^\top \mH+\mH^\top \bar{\xi}_\mH\right](\mH^\top \mH)^{-1}
    \\
    &-\mG(\bar{\xi}_{\mH}^T \bar{\xi}_{\mH})(\mH^\top \mH)^{-1}.
    \end{split}
\end{align}
Proof in \Cref{ap:proofs_attention_geodesic}.




\vspace{-0.2cm}
\section{Results}
\subsection{Diagonal Networks}

\paragraph{Experimental setup:}
We emulate the setup used in \citet{andriushchenko2023modern}. We generate a randomly distributed data matrix $\Xm$, a random ground-truth vector $\bm{\beta}^*$ that is $90\%$ sparse, and we train 50 diagonal networks to $10^{-5}$ training loss on a regression task. We take $d = 200$.

\begin{figure}[h!]
        \centering
        \includegraphics[trim=2cm 0.2cm 2cm 1cm,clip=true,width=0.5\textwidth]{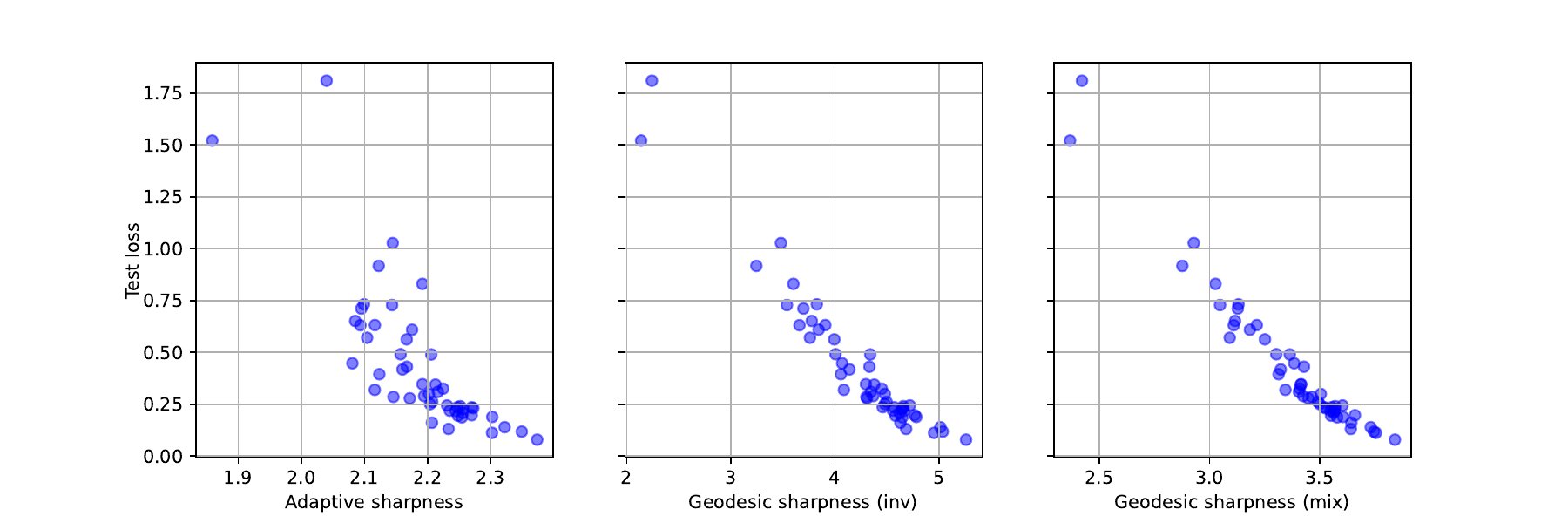}
        \caption{
          We show for 50 diagonal models trained on sparse data the generalization gap (here just the test loss) vs.
          worst-case adaptive sharpness (left, $\tau=-0.68$), $\langle \cdot, \cdot \rangle^{\text{inv}}$-geodesic sharpness (middle, $\tau=-0.83$), and $\langle \cdot, \cdot \rangle^{\text{mix}}$-geodesic sharpness (right, $\tau=-0.86$).
          The x axis is the value of the sharpness measure being considered, and the y axis is the test error.
        }\label{fig:diagonal}
\end{figure}

\paragraph{Results:} All three notions of sharpness are able to predict, to some degree, generalization (\cref{fig:diagonal}). Geodesic sharpness, although closely related for diagonal nets to adaptive worst-case sharpness, does slightly better. This applies to both metrics studied, and they perform roughly the same.
\vspace{-0.3cm}

\subsection{Transformers}

Transformers have a mix of attention layers and layers with more restricted
symmetries for which adaptive sharpness is more appropriate. We present in
Appendix~\ref{ap:schema} more details on how we treat the multi-layer schema of
transformers. In Appendix~\ref{ap:algorithm} we present
Algorithm~\ref{ap:alg_apgd}, which we use to solve for geodesic sharpness.

\vspace{-0.3cm}

\subsubsection{Vision Transformers}
\paragraph{Experimental setup:} We follow \citet{andriushchenko2023modern}, and look at models obtained from fine-tuning CLIP on ImageNet-1k~\citep{clip}.
To be more specific, we study the already trained classifiers obtained by \cite{wortsman2022soups}, after fine-tuning a CLIP ViT-B/32 on ImageNet with randomly selected hyperparameters.
We evaluate our measure and adaptive worst-case sharpness on the same 2048 data points from the training set of ImageNet, divided into batches of 256 points.
We calculate sharpness on each batch separately and average the results.
We take the generalization gap to be the difference between test and training error.

\begin{figure}[h!]
  \centering
  \includegraphics[trim=2cm 0.2cm 2cm
  1cm,clip=true,width=0.5\textwidth]{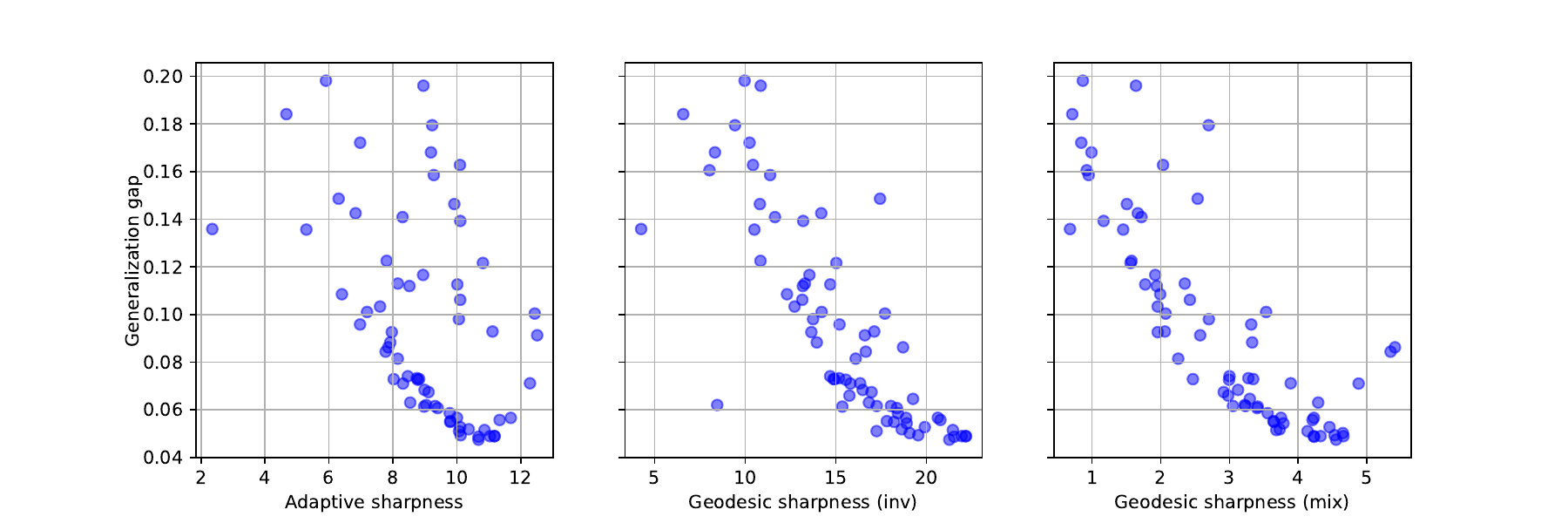}
  \caption{ We show for the
    72 models from \cite{wortsman2022soups} the generalization gap on ImageNet
    vs. worst-case adaptive sharpness (left, $\tau=-0.41$), $\langle \cdot,
    \cdot \rangle^{\text{inv}}$-geodesic sharpness (middle, $\tau=-0.71$), and
    $\langle \cdot, \cdot \rangle^{\text{mix}}$-geodesic sharpness (right,
    $\tau=-0.70$).}\label{fig:transformer_results}
\end{figure}

\vspace{-0.3cm}
\paragraph{Results:} In Figure~\ref{fig:transformer_results} we show our
results. We find a strong correlation between geodesic sharpness and the
generalization gap on ImageNet. This correlation is stronger than that observed
with adaptive sharpness and is consistently negative, implying that the
geodesically sharpest models studied on ImageNet are those that generalize best,
contrary to what might have been expected, but consistent with the correlation from the diagonal networks.

\vspace{-0.3cm}

\subsubsection{Language models}\label{sec:bert}
\paragraph{Experimental Setup:} We also validate our measure on 35 models from
~\citet{mccoy2020bertsfeather}, obtained after fine-tuning BERT on
MNLI~\citep{mnli2018}. We evaluate our measure and adaptive worst-case sharpness
on the same 1024 data points from the MNLI training set, with batches of 128
points. As done for ImageNet, we calculate sharpness on each batch separately
and average the results. \vspace{-0.1cm}
\paragraph{Results:}
\begin{figure}[h!]
  \centering
  \includegraphics[trim=2cm 0.2cm 2cm
  1cm,clip=true,width=0.5\textwidth]{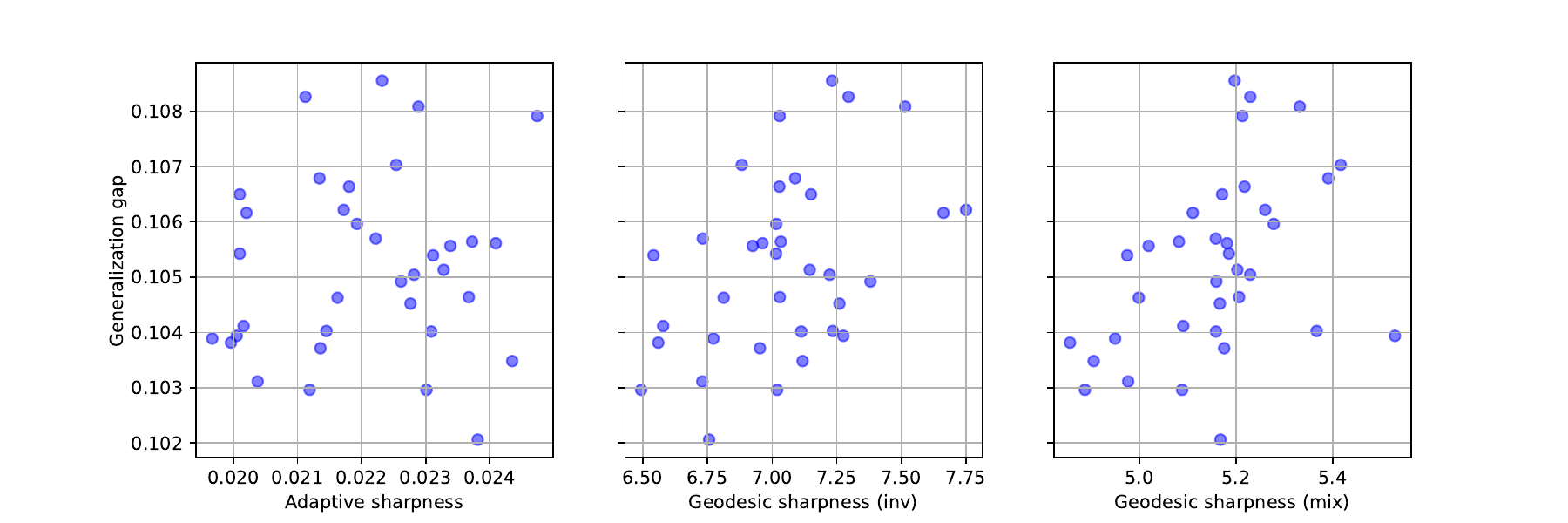}
  \caption{ We show for 35
    models from~\citep{mccoy2020bertsfeather} the generalization gap on the MNLI
    dev matched set \citep{mnli2018} vs. worst-case adaptive sharpness (left,
    $\tau=0.06$), $\langle \cdot, \cdot \rangle^{\text{inv}}$-geodesic sharpness
    (middle, $\tau=0.28$), and $\langle \cdot, \cdot
    \rangle^{\text{mix}}$-geodesic sharpness (right,
    $\tau=0.38$). \label{fig:bert}}
\end{figure}
We show our results in Figure~\ref{fig:bert}. We find a consistent correlation
between geodesic sharpness and the generalization gap on MNLI, for both metrics,
while adaptive sharpness, with $\tau = 0.06$, is unable to find any correlation.
The correlation is positive, indicating that the geodesically flattest models
generalize best. \vspace{-0.2cm}

\section{Remarks, Limitations \& Future Work}

We summarize all correlation coefficient results in~\cref{tab:results}.
The results of our ablation studies are in~\cref{ap:ablations}.

\vspace{-0.2cm}
\paragraph{Discovering correlation: } \hspace{-0.4cm} Adaptive sharpness, as discussed thoroughly by \citet{andriushchenko2023modern}, breaks down for transformers, and is unable to recover clear correlation between sharpness and generalization.
In contrast, geodesic sharpness consistently recovers strong correlation in the case of transformers, and consistently strengthens the correlation in the case of diagonal networks.
\vspace{-0.2cm}
\paragraph{Metric choice: } Our results are robust w.r.t.\,the choice of metric, as long as {\it it captures the parameter symmetry}.
The mixed metric yields better results on BERT, perhaps owing to its superior numerics (e.g.
possible inversion of nearly singular matrices is side-stepped), but not by much.
Additionally, the mixed metric avoids calling expensive Sylvester equation solvers and has a simple projection onto the horizontal space.
\vspace{-0.45cm}
\paragraph{Sign of the correlation: } One of the surprising results of this work is that the sign of the correlation between geodesic sharpness and generalization varies depending on setting and is at times negative, somewhat at odds with the common view that sharpness ``always'' correlates with generalization gap (i.e., flatter models generalize better).
This is not inherent to our proposed metrics: indeed, adaptive sharpness (previously found to positively correlate with generalization on other tasks~\citep{kwon2021asam}), anti-correlates with generalization in our diagonal network setting.
But where adaptive sharpness finds no correlation, our metrics do find a signed correlation, and where adaptive sharpness finds signed correlation, our metrics find a stronger similarly-signed correlation.
That is, we for the first time observe {\it consistent correlations} within-task for transformers, opening new questions for further study, as we discuss below.

\paragraph{Limitations:} While our \textit{geodesic sharpness} is more general than previous measures, there remain symmetries for which taking the quotient may be computationally expensive or intractable.
Still, it provides an important tool for analysing NNs: accounting for some symmetry is better than none, and under computational constraints it could be useful as a diagnostic ``probe''.


We set out to find an improved measure of sharpness for predicting generalization.
Our new measures detect previously undetected correlation with generalization.
In the process, however, we also discovered that the sign of the correlation, while consistent across metrics and models, can vary across tasks and/or data.
Until this new variability is understood, this limits its utility as a predictive tool.

While even observing this correlation is encouraging, not having a full picture of how this correlation behaves limits the utility of our new method in optimization, for example.

\paragraph{Future work}

Our work sets the stage to explore a new line of questioning: what is the role of data and how can it be integrated into our overall framework?
A more complete understanding of the simultaneous dual invariance induced by data and parameter symmetries is of interest and might help explain when geodesic sharpness correlates or anti-correlates with generalization.
As demonstrated by~\citet{foret2021sharpnessaware}, having a relatively simple quantity accessible during training that correlates to generalization is useful; further work may enable creation of similar optimizers better suited for large scale transformers.


\section{Conclusion}
\vspace{-0.1cm} In this paper, motivated by the success of adaptive sharpness measures in the study of generalization, we propose a novel adaptive sharpness measure: geodesic sharpness. We frame it in the context of Riemannian geometry and provide 
an effective method for including various parameter symmetries in the calculation of sharpness.
When the curvature introduced by parameter symmetry is disregarded, we find geodesic sharpness reduces to traditional adaptive sharpness measures.
We analytically investigate our measure on widely studied diagonal networks and empirically verify our approach on large scale transformers, finding a strong correlation between geodesic sharpness and generalization.


\newpage
\section*{Impact Statement}
\vspace{-0.1cm}
This paper presents work whose goal is to advance the study of deep learning. There are potential indirect societal consequences of our work, none which we feel must be specifically highlighted here.





\section*{Acknowledgements}


We would like to express our sincere gratitude to Agustinus Kristiadi and Rob Brekelmans for early feedback on the manuscript.
Resources used in preparing this research were provided, in part, by NSERC, the Province of Ontario, the Government of Canada through CIFAR, and companies sponsoring the Vector Institute www.vectorinstitute.ai/\#partners.


\medskip

{
  \small
  \bibliographystyle{icml2025.bst} 
  \bibliography{references.bib}
}

\clearpage
\appendix
\onecolumn
\thispagestyle{plain}
{\center\baselineskip 18pt
  \toptitlebar{\Large\bf \papertitle \ (Supplemental Material)}\bottomtitlebar
}
We provide in Table 1 a summary of correlation coefficients between sharpness and generalization for our experiments.
\begin{table}[h!]
    \centering
    \begin{tabular}{*4c}
        \toprule
        & \multicolumn{3}{c}{Rank correlation coefficient $\tau$} \\
        \cmidrule(lr){2-4}
        Setting & Adaptive sharpness & $\langle \cdot, \cdot \rangle^{\text{inv}}$ - geodesic sharpness & $\langle \cdot, \cdot \rangle^{\text{mix}}$- geodesic sharpness ()\\
        \midrule
        Diagonal networks &     -0.68   & -0.83 & \textbf{-0.86} \\
        ImageNet &    -0.41     &    \textbf{-0.71}   & -0.7 \\
        MNLI &       0.06 &  0.28  & \textbf{0.38}\\
        \bottomrule
    \end{tabular}
        \caption{Summary of the correlation between sharpness measures and generalization. We boldface the best performing metric}\label{tab:results}
\end{table}

In the sections that follow, we provide additional details to supplement the main text.

\section{Additional Examples of GL symmetries Symmetries in Neural Networks}
\begin{example}[Self-attention~\citep{vaswani2017attention}]\label{ex:self-attention}
  Given a sequence $\mX \in \sR^{t \times d}$ with $t$ tokens and model dimension $d$, self-attention (SA) uses four matrices $\mW_{\text{q}}, \mW_\text{k} \in \sR^{d \times d_{\text{k}}}, \mW_{\text{v}}, \mW_{\text{o}}^{\top} \in \sR^{d \times d_{\text{v}}}$ (usually, $d = d_{\text{v}} = d_{\text{k}}$) to produce a new $t \times d$ sequence
  \begin{align}\label{eq:self-attention}
    \begin{split}
      &\mathrm{SA}(\mW_{\text{q}}, \mW_\text{k}, \mW_\text{v}, \mW_{\text{o}})
      \\
      &=
        \mathrm{softmax}
        \left(
        \frac{
        \mX \mW_{\text{q}} \mW_\text{k}^{\top} \mX^{\top}
        }{\sqrt{d_{\text{k}}}}
        \right)
        \mX \mW_\text{v} \mW_{\text{o}}\,.
    \end{split}
  \end{align}
  This block contains two $\GL$ symmetries:
  one of dimension $d_{\text{k}}$ between the key and query projection weights, $\mG, \mH \leftarrow \mW_{\text{q}}, \mW_{\text{k}}$, and one of dimension $d_{\text{v}}$ between the value and out projection weights, $\mG, \mH \leftarrow \mW_{\text{v}}, \mW_{\text{o}}^{\top}$.
  Similar to Eq. \ref{eq:including-bias-terms}, we can account for biases in the key, query, and value projections by appending them to their weight,
  \begin{align*}
    \mG, \mH
    \leftarrow
    \begin{pmatrix}
      \mW_{\text{k}} \\ \vb_{\text{k}}^{\top}
    \end{pmatrix},
    \begin{pmatrix}
      \mW_{\text{q}} \\ \vb_{\text{q}}
    \end{pmatrix}^{\top}\,,
    \quad
    \mG, \mH
    \leftarrow
    \begin{pmatrix} \mW_{\text{v}} \\ \vb_{\text{v}} \end{pmatrix},
    \mW_{\text{o}}^{\top}\,.
  \end{align*}
  Commonly, $H$ attention heads $\{\mW^i_{\text{q}}, \mW^i_{\text{k}}, \mW^i_{\text{v},i}, \mW^i_{\text{o}} \}_{i=1}^H$ independently process $\mX$ and concatenate their results into the final output (usually $d_{\text{k}} = d_{\text{v}} = \nicefrac{d}{H}$).
  This introduces $2H$ $\GL$ symmetries.
  Everything also applies to general attention where, instead of $\mX$, independent data is fed as keys, queries, and values to Eq. \ref{eq:self-attention}.
\end{example}

\begin{example}[Shallow linear net]\label{ap:ex:shallow-linear-net}
  Consider a two-layer linear net $\mathrm{NN}(\mW_2, \mW_1) = \mW_2 \mW_1 \vx$ with weight matrices $\mW_1 \in \sR^{h \times d_{\text{in}}}, \mW_2 \in \sR^{d_{\text{out}} \times h}$ and some input $\vx \in \sR^{d_{\text{in}}}$.
  This net has $\GL$ symmetry with correspondence $\mG, \mH \leftarrow \mW_2, {\mW_1}^{\top}$ to \Cref{def:building-block-gl-symmetry}.
  With first-layer bias, we have
  \begin{align}\label{eq:including-bias-terms}
    \mW_2 (\mW_1 \vx + \vb_1)
    =
    \mW_2
    \begin{pmatrix}
      \mW_1 & \vb_1
    \end{pmatrix}
    \begin{pmatrix}
      \vx
      \\
      1
    \end{pmatrix}\,,
  \end{align}
  corresponding to $\mG, \mH \leftarrow \mW_2, \begin{pmatrix} \mW_1 & \vb_1 \end{pmatrix}^{\top}$. 
\end{example}

\begin{example}[Low-rank adapters (LoRA, \citet{hu2022lora})]\label{ap:ex:lora}
  Fine-tuning tasks with large language models add a trainable low-rank perturbation $\mL \in \sR^{d_1 \times h}, \mR \in \sR^{d_2 \times h}$ to the pre-trained weight $\mW \in \sR^{d_1 \times d_2}$,
  \begin{align}\label{eq:lora}
    \mathrm{LoRA}(\mW) = \mW + \mL \mR^{\top}\,,
  \end{align}
  introducing a $\GL(h)$ symmetry where $\mG, \mH \leftarrow \mL, \mR$.  \citet{lora-rite} propose an invariant way to train the parameters $\mL, \mR$ and show that doing so improves the result obtained via LoRA.
\end{example}
\section{Concepts and Review for Riemannian Geometry}
\label{ap:mathdefs}

Recall that $\overline{\gM}$ is the total space: the manifold of parameters of our network. Also, on the total space we have a loss function $\ell: \overline{\gM} \rightarrow \mathbb{R}$. Useful resources are~\citet{lee2003introduction},~\citet{absil2008optimization}, and~\citet{boumal2023introduction}.

\subsection{Orbit of $x$}
\label{ap:mathdefs:equiv}
A symmetry relation naturally defines an equivalence relation: two points $x, y \in \overline{\gM}$ are equivalent under the symmetry, if they can be mapped onto each other by the action,
\begin{align}
	x \sim y
	\quad \Leftrightarrow \quad
	\exists g \in \gG: y = \psi(g, x)\,.
\end{align}
In other words, if we let $\orbit(x) \coloneq \{ \psi(g, x) \mid g \in \gG \}$ be all points on the total space that are reachable from $x$ through the action of $\gG$, all points in an orbit are equivalent.
Instead of $\orbit(x)$, we will write
\begin{align}
	[x] \coloneqq \{ y \in \overline{\gM} \mid y \sim x \}
\end{align}
for the symmetry-induced equivalence class $[x]$ of $x \in \overline{\gM}$.

Let's further assume that $\overline{\ell}$ is symmetric under $\gG$, i.e.\,for any $x \in \overline{\gM}$ and all $g \in \gG$, $\overline{\ell}(x) = \overline{\ell}(\psi(g, x))$.

\subsection{Quotient $\gM$ and Natural Projection}
\label{ap:mathdefs:quotient}
If we take the quotient of the original parameter space manifold $\overline{{\gM}}$, by the equivalence relation, $\sim$, induced by the symmetries of our neural architecture, we get a quotient $\gM = \overline{\gM}/\sim$.  Under certain conditions, $\gM$ is a quotient manifold. The mapping between a point in total space to its equivalence class is called the natural projection:
\begin{definition}
	Let $\pi: \overline{\mathcal{M}} \rightarrow  \overline{\mathcal{M}}/\sim$, be defined by $ \overline{x} \mapsto x$. $\pi$ is called the natural, or canonical projection. We use $\pi(\overline{x})$ to denote $x$ viewed as a point of $\gM \coloneq \overline{\mathcal{M}}/\sim$.
\end{definition}

\subsection{Tangent Space, Vertical and Horizontal Spaces}
\label{ap:mathdefs:tangents}

Tangent vectors on the total space $\overline{\gM}$, embedded in a vector space $\mathcal{E}$ can be viewed as tangent vectors to $\mathcal{E}$, but the tangent space of the quotient manifold, $\gM$ is not as straightforward. 
First, note that any element  $\bar{\xi} \in T_{\bar{x}} \mathcal{M}$ that satisfies $D \pi(\bar{x}) [\bar{\xi}] = \xi$ (where $D$ is the differential) is a candidate for a representation of $\xi  \in T_{x} \mathcal{M}$. These aren't unique, and as we wish to work without any numerical ambiguity we introduce the notions of the vertical and horizontal spaces:

\begin{definition}
	For a quotient manifold $\mathcal{M} = \mathcal{M}/\sim$, the vertical space at $\bar{x} \in \mathcal{M}$ is the subspace $V_{\bar{x}} = T_{\bar{x}} \mathcal{F} = \ker D\pi(x)$
	where $\mathcal{F} = \{ \bar{y} \in \mathcal{M} : \bar{y} \sim \bar{x} \}$ is the fiber of $\bar{x}$. The complement of $V_{\bar{x}}$ is the horizontal space at $\bar{x}$:
	$T_{\bar{x}}\overline{\gM} = V_{\bar{x}} \oplus H_{\bar{x}}$.
\end{definition}

\begin{definition}
	There is only one element  $\bar{\xi}_{\bar{x}}$ that belongs to $H_{\bar{x}}$ and satisfies $D \pi(\bar{x}) [\bar{\xi}_{\bar{x}}] = \xi$. This unique vector is called the {\it horizontal lift} of of $\xi$ at $\bar{x}$. We denote the operator that affects the procedure by $\text{lift}_{\bar{x}}(\cdot)$ When the ambient space, $\mathcal{E}$ is a subset of $\mathbb{R}^{n \times p}$, the horizontal space can also be seen as such a subset, providing a convenient matrix representation of {\it a priori} abstract tangent vectors of $\gM$.
\end{definition}

\subsection{Riemannian Manifold}
\label{ap:mathdefs:riemannian}

We give our total space $\overline{\gM}$ a smooth inner product over its tangent vectors to give a Riemannian manifold. 

\begin{definition}
	A Riemannian manifold is a pair ($\mathcal{M}$, $g$), where $\mathcal{M}$ is a smooth manifold and $g$ is a Riemannian metric, defined as the inner product on the tangent space $T_x \mathcal{M}$ for each point $x \in \mathcal{M}$, $g_x(\cdot, \cdot): T_x \mathcal{M} \times T_x \mathcal{M} \rightarrow \mathbb{R}$. We also use the notation $\langle \cdot, \cdot \rangle_x$ to denote the inner product.
\end{definition}

Note that this definition is not as arcane as it may appear since any smooth manifold admits a Riemannian metric, and we can consider the space of parameters of most neural architectures as constituting a smooth manifold, admitting at least a simple, Euclidean, metric. 

The horizontal space can now be defined as the {\it orthogonal} complement of the vertical space: $H_{\bar{x}} = (V_{\bar{x}})^\perp = \{ u \in T_{\bar{x}} \overline{\mathcal{M}} : \langle u, v \rangle_x = 0 \text{ for all } v \in V_{\bar{x}} \}$. Additionally, letting $\bar{g}_{\bar{x}}$ denote the metric on $\overline{\gM}$, if for every $x \in \gM$ and every $\xi_x,\zeta_x$ in $T_x \gM$, $\bar{g}_{\bar{x}}(\bar{\xi}_{\bar{x}},\bar{\zeta}_{\bar{x}})$ does not depend on $\bar{x} \in \pi^{-1}(x)$ then, $g_{x}(\xi_{x},\zeta_{x})=\bar{g}_{\bar{x}}(\bar{\xi}_{\bar{x}},\bar{\zeta}_{\bar{x}})$ defines a valid metric on the quotient manifold $\gM$.

\subsection{Riemannian Gradient}
\label{ap:mathdefs:rgrad}

\begin{definition}
	If $\bar{f}$ is a smooth scalar field on a Riemannian manifold $\overline{\gM}$, then the {\it gradient} of $\bar{f}$ at $\bar{x}$, $\text{grad} \bar{f}(\bar{x})$ is the unique element of $T_{\bar{x}} \overline{\gM}$ such that
	\begin{equation*}
		\langle \text{grad} \bar{f}(\bar{x}), \bar{\xi} \rangle_{\bar{x}} = D \bar{f}(\bar{x})[\bar{\xi}], \forall \bar{\xi} \in T_{\bar{x}} \overline{\gM}
	\end{equation*}
	If $\bar{f}$ is a function on $\overline{\gM}$, that induces a function $f$ on a quotient manifold $\gM$ of $\overline{\gM}$, then we can express the horizontal lift of $\text{grad } f$ at $\bar{x}$ as
	\begin{equation*}
		\text{lift}_{\bar{x}}(\text{grad f}) = \text{grad} \bar{f}(\bar{x}).
	\end{equation*}
\end{definition}

\subsection{Geodesic Curves}
\label{ap:mathdefs:geodesics}
\begin{definition}
   \leavevmode\vspace{0.5\baselineskip}
	\begin{itemize}
		\item[(a)]	Geodesic curves, $\bar{\gamma}$, are the curves of minimal distance between two points on a manifold $\overline{\gM}$. The distance along the geodesic is called the {\it geodesic distance}. If $\gM$ is a Riemannian quotient manifold of $\overline{\gM}$, with canonical projection $\pi$, and $\bar{\gamma}$ is a geodesic on $\overline{\gM}$, then $\gamma = \pi \circ \bar{\gamma}$ is a geodesic curve on $\gM$. 
		\item[(b)] Alternatively, geodesics, $\bar{\gamma}(t)=0$ can be defined as curves from a given point $p \in \overline{\gM}$, (i.e., $\bar{\gamma}(0) = p$), with initial {\it velocity}, $\dot{\bar{\gamma}}(0) = \bar{\xi} \in T_{\bar{p}}\overline{\gM}$, such that their {\it acceleration} is zero (a generalization of Euclidean straight lines). This characterization provides us with the following equation in local coordinates for the geodesic:
  \begin{align*}
      \frac{d^2 \gamma^{\lambda}}{dt^2} + \Gamma^\lambda_{\mu\nu} \frac{d\gamma^\mu}{dt} \frac{d\gamma^\nu}{dt} = 0
  \end{align*}where $\Gamma^\lambda_{\mu\nu}$ are the Christoffel symbols, $\Gamma^\lambda_{\mu\nu} = \frac{1}{2} g^{\lambda\sigma} \left( \frac{\partial g_{\sigma\mu}}{\partial x^\nu} + \frac{\partial g_{\sigma\nu}}{\partial x^\mu} - \frac{\partial g_{\mu\nu}}{\partial x^\sigma} \right)$. Additionally, the geodesics can also be derived as the curves that are minima of the energy functional
  \begin{align*}
      S(\gamma) = \int_a^b g_{\gamma(t)} (\dot{\gamma(t)},\dot{\gamma(t)}) dt
  \end{align*}
  This second perspective will prove useful for the geodesics of the attention layers.
 
If the initial velocity tangent vector, $\xi$, is horizontal then, $\forall t, \dot{\bar{\gamma}}(t) \in H_{\bar{\gamma}(t)}$, that is, if the velocity vector starts out as horizontal, then it will stay horizontal. We call these geodesics, {\it horizontal geodesics}. The curve $\gamma = \pi \circ \bar{\gamma} $ is a geodesic of the quotient manifold $\gM$, with the same length as $\bar{\gamma}$. This also holds the other way, i.e., a geodesic in the quotient manifold can be lifted to a horizontal geodesic in the total space.
	\end{itemize}
\end{definition}
\section{Geodesic sharpness: practical concerns}\label{ap:optimization}
\subsection{Transformers}\label{ap:schema}
Transformers, introduced by \citet{vaswani2017attention}, consist of multiheaded self-attention and feedforward layers, both wrapped in residual connections and layer normalizations. Visual transformers, in addition, tend to have convolutional layers. 

Mathematically, focusing for the moment on the multi-headed attention blocks,
\begin{align*}
\text{MultiHead}(Q, K, V) &= \big[\text{head}_1, \dots, \text{head}_h\big] W^o \\
\text{where} \quad \text{head}_i &= \text{Attention}\big(QW_i^Q, KW_i^K, VW_i^V\big)
\end{align*}
where ${\text{Attention}}(Q, K, V) = \text{softmax}\left(\frac{QK^{T}}{\sqrt{d_k}}\right)V$. 
From this we can ascertain the following symmetries:
\begin{align*}
&1)~~~(W^Q_i, W^K_i) \rightarrow (W^Q_i G^{-1},W^K_i G^T)~, \forall G \in \text{GL}_n(d_{\text{head}})
\\
&2)~~~(W^V_i, W^o_i) \rightarrow (W^V_i G^{-1},W^o_i G^T)~, \forall G \in \text{GL}_n(d_{\text{head}}) 
\end{align*}
where $W^o_i$ are the columns of $W^o$ that are relevant for the matrix multiplication with each $W^V_i$, taking into consideration the head concatenation procedure.

In the full transformer model when solving for geodesic sharpness, for each layer, we apply Eq.~\ref{eq:geodesic_approximation} to each $(W^Q_i, W^K_i)$ and $(W^V_i, W^o_i)$, using Eq.~\ref{eq:attention_approximation}. This results in horizontal vectors $(\bar{\xi}^Q_i,\bar{\xi}^K_i)$ and $(\bar{\xi}^V_i,\bar{\xi}^o_i)$. 
For the non-attention parameters, $\vw$, (belonging to fully connected layers, convolutional layers and layer norm), we keep to the recipe of adaptive sharpness, so that $||\bar{\xi}_{\vw}|| = ||\left(\bar{\xi}_{\vw} \odot |\vw|^{-1} \right)||_2$. The norm of the full update vector, $\bar{\xi} = \text{concat}(\bar{\xi}^Q_i,\bar{\xi}^K_i,\bar{\xi}^V_i,\bar{\xi}^o_i,\bar{\xi}_{\vw})$, where a sum over all parameters of the network is implicit, is $||\bar{\xi}||^2 = \sum \left( ||(\bar{\xi}^Q_i,\bar{\xi}^K_i)||^2+||(\bar{\xi}^V_i,\bar{\xi}^o_i)||^2+||\bar{\xi}_{\vw}||^2 \right)$.
\subsection{Algorithm}\label{ap:algorithm}
Following the lead of \citet{andriushchenko2023modern}, we use Auto-PDG, proposed in \citet{croce2020reliable}, but now optimizing the horizontal vector $\bar{\xi}$ instead of the input. In Algorithm~\ref{ap:alg_apgd}, $\ell$ is the loss over the batch we are optimizing over, S is the feasible set of horizontal vectors, $\bar{\xi}$, with norm smaller than $\rho$, and $P_S$ is the projection onto this set. $\Gamma$ are the Christoffel symbols for the parameters. $\eta$ and $W$ are fixed hyperparameters, which we keep as in \citet{andriushchenko2023modern}, and the two conditions in Line~\ref{alg:condition} can be found in \citet{croce2020reliable}. The only differences to the algorithm employed to calculate adaptive sharpness are in lines~\ref{alg:line3},~\ref{alg:line8},~\ref{alg:line9}, and~\ref{alg:line10}. For the metric $\langle\cdot, \cdot \rangle^{\text{mix}}$ the only differences are in the Christoffel symbols and in the Riemannian gradient ($\nabla_{\mG} \ell \rightarrow \nabla_{\mG} \ell \left(\mH^T \mH \right)^{-1}$)

\begin{algorithm}[h]
 \caption{Auto-PGD}\label{ap:alg_apgd}
    \begin{algorithmic}[1]
        \State {\bfseries Input:} objective function $\ell$, perturbation set $S$, $\bar{\xi}^{(0)}$, initial weights $\iter{\vw}{0}$, $\eta$, $N_\textrm{iter}$, $W=\{w_0, \ldots, w_n\}$ %
        
        \State {\bfseries Output:} $\bar{\xi}_\textrm{max}$, $\ell_\textrm{max}$
        
        \State $\iter{\vv}{1} \gets \vw^{(0)}+\bar{\xi}^{(0)}-\frac{1}{2}\Gamma\iter{\bar{\xi}}{0}\iter{\bar{\xi}}{0}$ \Comment{Perturb weights according to Eq.~\ref{eq:geodesic_approximation}}\label{alg:line3}
        
        \State $\iter{\bar{\xi}}{1} \gets P_\mathcal{S}\left(\iter{\bar{\xi}}{0} + \eta \nabla_{\bar{\xi}} \ell(\iter{\vv}{1})\right)$
        
        \State $\ell_\textrm{max}\gets \max\{\ell(\iter{\vw}{0}), \ell(\iter{\vv}{1})\}$
        
        \State $\bar{\xi}_\textrm{max} \gets \iter{\bar{\xi}}{0}$ \textbf{if} $\ell_\textrm{max}\equiv \ell(\iter{\vw}{0})$ \textbf{else} $\bar{\xi}_\textrm{max} \gets \iter{\bar{\xi}}{1}$ %
        
        \For{$k=1$ {\bfseries to}  $N_\textrm{iter}- 1$}
        
        \State $\iter{\vv}{k+1} \gets \vw^{(0)}+\bar{\xi}^{(k)}-\frac{1}{2}\Gamma\iter{\bar{\xi}}{k}\iter{\bar{\xi}}{k}$ \Comment{Perturb weights according to Eq.~\ref{eq:geodesic_approximation}} \label{alg:line8}
        
        \If{$\vw^{(0)}$ is an attention weight}
        \State $g \gets  \nabla_{\bar{\xi}} \ell(\iter{\vv}{k+1})\vw^{(0), T} \vw^{(0)}$ \Comment{Make attention gradients Riemannian}\label{alg:line9}
        
        \Else
        \State $g \gets  \nabla_{\bar{\xi}} \ell(\iter{\vv}{k+1})\odot(\vw^{(0)})^2$ \Comment{Make the other gradients Riemannian}\label{alg:line10}
        \EndIf
        \State $\iter{\vz}{k+1} \gets P_\mathcal{S}\left(\iter{\bar{\xi}}{k} + \eta g)\right)$

        \State $\iter{\bar{\xi}}{k+1} \gets  P_\mathcal{S} \left(\iter{\bar{\xi}}{k} + \alpha(\iter{\vz}{k+1}- \iter{\bar{\xi}}{k})  + (1-\alpha) (\iter{\bar{\xi}}{k} - \iter{\bar{\xi}}{k-1})\right)$
        
        \If{$\ell(\iter{\vv}{k+1}) > \ell_\textrm{max}$}
        
        \State $\bar{\xi}_\textrm{max} \gets \iter{\bar{\xi}}{k+1}$ and $\ell_\textrm{max}\gets \ell(\iter{\vv}{k+1})$
        \EndIf
        
        \If{$k\in W$}
        
        \If{Condition 1 {\bfseries or}
        Condition 2}\label{alg:condition}
        
        \State $\eta \gets \eta / 2$ and  $\iter{\vw}{k+1} \gets \vw_\textrm{max}$
        \EndIf
        \EndIf
        \EndFor
    \end{algorithmic}
\end{algorithm}

\subsection{Complexity}

Geodesic sharpness is slightly more expensive than adaptive sharpness in the following sense: Our approach consists of three steps: 1) perturbing the weights according to Eq.~\ref{eq:geodesic_approximation}, 2) optimizing the perturbations with gradient descent, and 3) projecting them onto the feasible set, i.e. horizontal vectors within the geodesic ball with a small enough norm.

Steps 1) and 2) are also present in adaptive sharpness. Step 1) in our approach is slightly more expensive because we need to evaluate the quadratic form that involves the Christoffel symbols (given by Eq.~\ref{eq:attention_approximation} and Eq.~\ref{eq:attention_approximation_mix}); this step introduces $n_{\text{params}}$ weight matrix multiplications, but these are quite efficient. Making the gradients Riemannian, costs another $n_{\text{params}}$ weight matrix multiplications. Neither of these bottleneck our approach.
For $\langle \cdot, \cdot \rangle^{\text{inv}}$, Step 3) requires solving a Sylvester equation to project the direction of the updated geodesic back onto the horizontal space. This solve is cubic in $h$~\citep{sylvester2001}, but $h$ is usually small (e.g. $h = 64$ in the ImageNet and BERT experiments). For $\langle \cdot, \cdot \rangle^{\text{mix}}$, only efficient matrix multiplications are required.

On practical transformers, we expect the bottleneck to be the forward and backward propagations, just like in adaptive sharpness. 

\vspace{-0.5cm}
\section{Geodesic sharpness: Scalar Toy model}\label{ap:scalar_toy}
To make our approach explicit, we illustrate it on a NN with two scalar parameters $G$ and $H$, square loss, and a single (scalar) training point $(x,y)$. We use $\langle \cdot,\cdot \rangle^{\text{inv}}$ throughout. For this example, everything is analytically tractable. We also contrast our sharpness measure with previously proposed ones to highlight its invariance.

Since we require full column-rank, our parameter space is $\gM = \sR_{*} \times \sR_{*}$ with $\sR_{*} = \sR \setminus \{0\}$.

\paragraph{Metric} At a point $(G, H) \in \gM$, for two tangent vectors $\eta = (\eta_G, \eta_H)$, $\nu = (\nu_G, \nu_H) \in T_{(G, H)}\gM$, we have 
\begin{align} 
\langle \eta,\nu \rangle^{\text{inv}}
= \frac{\eta_G \nu_G}{G^2} 
+
\frac{\eta_H \nu_H}{H^2}
=
\eta^\top 
\underbrace{
\begin{pmatrix}
    \frac{1}{G^2} & 0
    \\
    0 & \frac{1}{H^2}
\end{pmatrix}
}_{g_{kl}}
\nu
\end{align}
We denote the inverse metric by $g^{kl} = \begin{pmatrix} G^2 & 0 \\ 0 & H^2 \end{pmatrix}$

\paragraph{Horizontal space} $H_{(G,H)} = \{\left(\eta_G, \eta_H\right) \in T_{(G, H)} \gM ~~|~~\frac{\eta_G}{G}=\frac{\eta_H}{H}\}$

\paragraph{Geodesics} To compute the geodesics on the quotient space, we need the Christoffel symbols $\Gamma^i_{km}$.

Using a coordinate system $(p^1, p^2)=(G,H)$, we have the following equation for a geodesic $\gamma(t) = (\gamma_G(t),\gamma_H(t))$, with initial conditions $\gamma(0) = (G_0,H_0)$ and $\dot{\gamma}(0) = (\eta_{G_0},\eta_{H_0})$

\begin{equation*} \frac{d^2 \gamma_G}{dt^2}+\Gamma_{11}^1\left(\frac{d \gamma_G}{d t}\right)^2=0
\end{equation*}
and similarly for $H$ with $\Gamma^2_{22}$ instead of $\Gamma^{1}_{11}$.

The Christoffel symbols can be found using the metric, $g$, and its inverse. Using the Einstein notation and denoting the inverse of $g$ by the use of upper indices:
\begin{equation*}
{\Gamma^i}_{kl}
  = \frac{1}{2} g^{im} \left(\frac{\partial g_{mk}}{\partial x^l} + \frac{\partial g_{ml}}{\partial x^k} - \frac{\partial g_{kl}}{\partial x^m} \right)
\end{equation*}

Then
\begin{align*}
   {\Gamma^1}_{11}&=\frac{1}{2} g^{1 m}\left(\frac{\partial g_{m 1}}{\partial p^1}+\frac{\partial g_{m 1}}{\partial p^1}-\frac{\partial g_{k l}}{\partial p^m}\right)=-\frac{1}{G} 
   \\
   {\Gamma^2}_{22}&=-\frac{1}{H} 
\end{align*}
All other Christoffel symbols are 0. Our geodesic equations then become (we omit the derivation for H, which is identical but with $G \leftrightarrow H$)
\begin{equation*}
    \frac{d^2 \gamma_G}{dt^2}-\frac{1}{\gamma_G}\left(\frac{d \gamma_G}{d t}\right)^2=0
\end{equation*}

This ODE has the (unique) solution $\gamma_G(t) = A_G\exp(B_G t)$. Taking into account the initial conditions, $A_G = G_0, A_H = H_0$ and due to the definition of the horizontal space, $B_G=\frac{\eta_G}{G_0}$ and $B_H = \frac{\eta_H}{H_0}$, this becomes
\begin{equation*}
    \gamma(t) = \left(G_0 \exp(\frac{\eta_G}{G_0} t), H_0 \exp(\frac{\eta_H}{H_0} t) \right)
\end{equation*}
One important detail to note is that these geodesics are not complete, that is, not all two points can be connected by a geodesic. Points with different signs cannot be connected, which makes sense since we excluded the origin from the acceptable parameters and in 1D we need to cross it to connect points with differing signs. All points that lie in the same quadrant as $(G_0, H_0)$ can be connected through a geodesic.

Putting it all together
\begin{align}
\begin{split}
	S_{\text{max}}^{\rho}((G_0,H_0)) &=  \left[ \max\limits_{||B|| \leq \rho}x^2G_0^2H_0^2(\exp(4B)-1)-2yxG_0H_0(\exp(2B)-1) \right], 
 \end{split}
\end{align}
Letting $y_0 = G_0 H_0 x$, this becomes:
\begin{align}\label{eq:max-sharpness-1d}
\begin{split}
	S_{\text{max}}^{\rho}((G_0,H_0)) &=  \left[ \max\limits_{||B|| \leq \rho}y_0^2(\exp(4B)-1)-2yy_0(\exp(2B)-1) \right], 
 \end{split}
\end{align}
Since $\eta_H$ is completely determined by $\eta_G$ we can ignore the maximization over it. 

Since in practice we'll take $\rho \ll 1$, we Taylor expand to get

\begin{equation*}
    S_{\text{max}}^{\rho} \approx 4 \rho |y_0| |y-y_0|
\end{equation*}

This presents an issue when the residual, $|y-y_0|$, is zero, so we can also expand to second order, to get, when $|y-y_0| \approx 0$

\begin{equation*}
    S_{\text{max}}^{\rho} \propto  \rho^2 |y_0| |y-2y_0| = 2  \rho^2 y_0^2
\end{equation*}

This is, up to constants, just $||G\odot H||_2^2$. This is also invariant to $GL_1$ transformations, as expected.

Very close to the minimum we only capture (second-order in $\rho$) properties of the network, a bit further away from it we capture a (first-order in $\rho$) mix of data and network properties. 

\paragraph{Comparison with more traditional measures}
The local average and worst case Euclidean sharpness (at a minimum) are
\begin{align*}
    &S_{\text{avg}} = \Tr \nabla^2 L_S = G^2+H^2
    \\
    &S_{\text{max}} = \lambda_{\text{max}}( \nabla^2 L_S) = G^2+H^2
\end{align*}

Adaptive sharpness is defined as
\begin{align*}
S^{\rho}_{\text{avg}}(w, c) &=\mathbb{E}_{S \sim \mathbb{P}_m} \left[ L_S(w + \delta) - L_S(w) \right],
\quad \delta \sim \mathcal{N}(0, \rho^2 \text{diag}(c^2))
\\
S^{\rho}_{\text{max}}(w, c) &= \mathbb{E}_{S \sim \mathbb{P}_m} \left[ \max_{\|\delta \odot c^{-1}\|_p \leq \rho} L_S(w + \delta) - L_S(w) \right],
\end{align*}

By picking $c$ very carefully one can get
\begin{align*}
S^{\rho}_{\text{avg}}(w, c) = |GH|
\\
S^{\rho}_{\text{max}}(w, c) = |GH|
\end{align*}
By contrast, in our approach there is no need for careful hyperparameter choices

\paragraph{Geodesic flatness with more data points}
How does the geodesic flatness look like with more data points?

\begin{align*}
    L_S(G,H) = \frac{1}{n} \sum_{i=1}^n (GHx_i-y_i)^2
\end{align*}

which leads to (defining $y^0_i=GH x_i$):
\begin{align} \label{eq:multisample_full_geodesic}
    S_{\text{max}}^{\rho} = \max_B \frac{1}{n}\sum_{i=1}^n \left[ (y_i^0)^2\left(\exp(\frac{B}{|B|}2\sqrt{2} \rho)-1\right)-2 yy_i^0\left(\exp(\frac{B}{|B|}\sqrt{2}\rho)-1\right) \right]
\end{align}
Taylor expanding (in $\rho$) once more, we see that
\begin{align} \label{eq:multisample_geodesic}
    S_{\text{max}}^{\rho} \approx \max_B \frac{1}{n}\sum_{i=1}^n \left[ 2\sqrt{2} \rho \frac{B}{|B|}  y_i^0(y_i^0-y)+2 \rho^2 (y_i^0)^2 \right]
\end{align}

Which $B$ maximizes Eq.~\ref{eq:multisample_geodesic}, depends on the sign of $\sum_{i=1}^n \left[  y_i^0(y_i^0-y)\right]$: $B <0$ if the sum is negative, the reverse if the opposite is true.

\subsection{Traditional flatness}
\begin{figure}[h!]
  \centering
  \hspace*{-0.03\linewidth}
  \begin{subfigure}[t]{0.25\linewidth}
    \caption{Loss}
    \centering
    \includegraphics{fig/exp03_illustration_toy/square_loss.pdf}
  \end{subfigure}
\hspace*{0.25\linewidth}
  \begin{subfigure}[t]{0.2\linewidth}
    \centering
    \textbf{Euclidean}
    \caption{Gradient norm}\label{subfig:visualization-scalar-toy-euclidean-gradient1}
    \includegraphics[width=\linewidth]{fig/exp03_illustration_toy/square_loss_gradient.pdf}
\caption{Hessian trace}
\label{subfig:visualization-scalar-toy-euclidean-hessian}
\includegraphics[width=\linewidth]{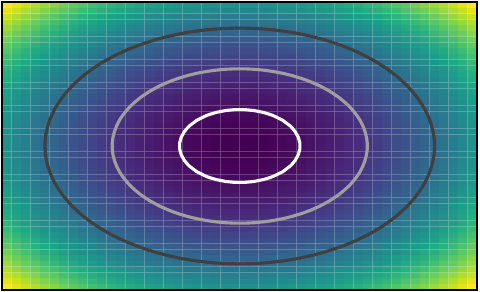}
  \end{subfigure}
  \begin{subfigure}[t]{0.2\linewidth}
    \centering
    \textbf{Riemannian}
    \caption{Gradient norm}\label{subfig:visualization-scalar-toy-riemannian-gradient1}
    \includegraphics[width=\linewidth]{fig/exp03_illustration_toy/square_loss_riemannian_gradient.pdf}
    \caption{Hessian trace}\label{subfig:visualization-scalar-toy-riemannian-hessian}
    \includegraphics[width=\linewidth]{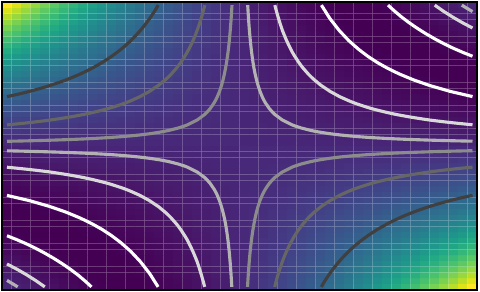}
  \end{subfigure}
  \caption{\textbf{Quantities from the Riemannian quotient manifold respect the loss landscape's symmetry; Euclidean quantities do not.}
    We use a synthetic least squares regression task with a two-layer NN $x \mapsto \theta_2 \theta_1 x$ with scalar parameters $\vtheta_i \in \sR$ and input $x \in \sR$.
    The NN is re-scale invariant, i.e.\,has $\GL(1)$ symmetry: For any $\alpha \in \sR \setminus \{ 0 \}$, the parameters $(\theta_1', \theta_2') = (\alpha^{-1}\theta_1, \alpha \theta_2)$ represent the same function.
    (\subref{subfig:visualization-scalar-toy-loss}) The loss function inherits this symmetry and has hyperbolic level sets.
    (\subref{subfig:visualization-scalar-toy-euclidean-gradient1},\subref{subfig:visualization-scalar-toy-euclidean-hessian}) The Euclidean gradient norm 
    does not share the loss function's geometry and changes throughout an orbit where the NN function remains constant.
    (\subref{subfig:visualization-scalar-toy-riemannian-gradient1},\subref{subfig:visualization-scalar-toy-riemannian-hessian}) The Riemannian gradient norm and Hessian trace 
    follow the loss function's symmetry and remain constant throughout an orbit, i.e.\,they do not suffer from ambiguities for two points in parameter space that represent the same NN function.
    All quantities were normalized to $[0; 1]$ and we fixed six points in parameter space and computed the level sets running through them to illustrate the geometry.
  }
  \label{fig:new}
\end{figure}

In Figure~\ref{fig:new} we extend Figure~\ref{fig:visualization-scalar-toy} to include the trace of the Hessian, both Euclidean and Riemannian. The trace of the network Hessian is a quantity that can be used to quantify flatness. We plot, for the scalar toy model, the level sets of: a) the loss function; b) the Euclidean and Riemannian gradient; c) the traces of the Euclidean and Riemannian network Hessian. Several features of the plots are important to note: a) the Riemannian version of the gradient and Hessian have the same level set geometry as the loss function; b) both the Riemannian gradient norm and the trace of the Riemannian Hessian have smaller values throughout than their Euclidean equivalents; c) the trace of the Riemannian Hessian actually reaches 0 when at the local minimum, whereas the Euclidean Hessian actually attains its highest value there; d) the Euclidean trace of the Hessian cannot distinguish between a minimum and a maximum whereas the Riemannian trace can actually do so.
Even for simple flatness measures, correcting for the quotient geometry can provide a much clearer picture.

\section{Geodesic sharpness: Diagonal networks in full generality}
\label{ap:diagonal}
\subsection{Metric (\ref{eq:ginv_diagonal})}
\paragraph{Metric:} At a point $(\uv, \vv) \in \gM$, for two tangent vectors $\eta = (\eta_{\uv}, \eta_{\vv})$, $\nu = (\nu_{\uv}, \nu_{\vv}) \in T_{(\uv, \vv)} \gM$, we have 
\begin{align}
\langle \eta, \nu \rangle^{\text{inv}}
= \sum_{i=1}^d \frac{\eta_{\uv}^i \nu_{\uv}^i}{(\uv^i)^2} +\frac{\eta_{\vv}^i \nu_{\vv}^i}{(\vv^i)^2}
\end{align}

\paragraph{Horizontal space:} $H_{(\uv, \vv)} = \{\left(\eta_{\uv}, \eta_{\vv} \right) \in T_{(\uv, \vv)} \gM ~~|~~\frac{\eta_{\uv}^i}{\uv^i}=\frac{\eta_{\vv}^i}{\vv^i} ~~~~~\forall i \in \{1,\ldots, d\}\}$
 \paragraph{Geodesics:} We define $\Bv^i= \frac{\eta_{\uv}^i}{\uv^i}=\frac{\eta_{\vv}^i}{\vv^i} \forall i \in \{1,\ldots, d\}$, so that

\begin{equation}\label{eq:diagonal_geodesic}
\bm{\gamma}(t)^i = (\uv(t), \vv(t)) = \left(\uv_0^i \exp(\Bv_i t), \vv_0^i \exp(\Bv_i t) \right) \forall i \in \{1,\ldots, d\}
\end{equation}

where $\uv_0^i$ and $\vv_0^i$ are the initial positions for our parameters, i.e., the parameters that the network actually learned.
\paragraph{Geodesic sharpness:} 

We assume that in what follows $\Xm^T\Xm = Id_d$, and we denote $\bv_0 = \uv_0 \odot \vv_0, \bm{\gamma}_t = \left(\exp(2\Bv^1 t), \ldots \exp(2\Bv^d t)\right), \bv_t = (\uv_t \odot \vv_t) = \bv_0 \odot \bm{\gamma}_t, \bv_* = \Xm^T y$. Note that $\bv_*$ is just the optimal least squares predictor when $\Xm^T\Xm = Id$.
With this notation
\begin{align} \label{eq:diagonal_geodesic_flatness}
S_{\text{max}} &= \max\limits_{||\Bv|| \leq \rho} \sum_i^d \left[(\bv_0^i)^2 (\bm{\gamma}_t \odot \bm{\gamma}_t-1)\right]-2(\bv_0 \odot \bm{\gamma}_t-1)^T \bv_*
\end{align}

At a first glance, this expression does not seem to have a simple interpretation, but we Taylor expand it to second order in $\Bv$ (since $\rho$ is supposed to be small):
\vspace{-0.2cm}
\begin{align} \label{eq:taylor_matrix_diagonal_geodesic_flatness}
S_{\text{max}} \approx  \max\limits_{||\Bv|| \leq \rho} 4 \Bv^T \vr+4 \Bv^T \Dv \Bv
\end{align}

where $\vr = \{\bv_0^i(\bv_0^i-\bv_*^i), i=1, \ldots, d\}$, $\vr^\prime = \{(\bv_0^i-\bv_*^i), i=1, \ldots, d\}$ and $\Dv = diag(\bv_0^i(2\bv_0^i-\bv_*^i)) = diag(\bv_0^i(\bv_0^i+(\vr^\prime )^i))$. We separate the analysis of Eq.\ref{eq:taylor_matrix_diagonal_geodesic_flatness} into three cases:

\subparagraph{case a): $\vr \neq 0$ and first order suffices}
Eq.\ref{eq:taylor_matrix_diagonal_geodesic_flatness} becomes
\vspace{-0.15cm}
\begin{align*}
	S_{\text{max}} = \max\limits_{||\Bv|| \leq \rho} 4 \Bv^T \vr
\end{align*}
with solution 
	$S_{\text{max}} = 4 \rho ||\vr||  $.
This is essentially the gradient norm-- a useful quantity for understanding generalization \citep{zhao2022penalizing}. 
\subparagraph{case b): $\vr = 0$}

Here we necessarily have to consider the second order terms, so that Eq.\ref{eq:taylor_matrix_diagonal_geodesic_flatness} becomes 
\begin{align*}
S_{\text{max}} = \max\limits_{||\Bv|| \leq \rho} 4 \Bv^T \Dv \Bv
\end{align*}
This has the well known solution of $S_{\text{max}} = \rho^2 \lambda_{\text{max}}(\Dv)= \rho^2 \max((\beta_0^i)^2)$. This is just $||\bm{\beta}||_{\infty}^2$, which is the square of what we would get by using adaptive sharpness, Eq.\ref{eq:max_def}, with a very carefully chosen hyper-parameter $\bm{c}$. This is a quantity that is useful when our ground-truth, $\bm{\beta}^*$ is dense. 

\subparagraph{case c): $\vr \neq 0$ and we need both first and second order terms}

In this case, Eq.\ref{eq:taylor_matrix_diagonal_geodesic_flatness} needs to be considered in full, and we solve the maximization problem using Lagrange multipliers. The Lagrangian will be
\begin{align*}
	\mathcal{L} = -4 \Bv^T \vr-4 \Bv^T \Dv \Bv+\lambda(\Bv^T \Bv-\rho^2)
\end{align*}

The KKT conditions then are

\begin{align} \label{eq:kkt_diagonal}
	&\frac{\partial \mathcal{L}}{\partial \Bv} = -4\vr-8 \Dv \Bv+2\lambda \Bv = 0
	\\
	&\lambda (\Bv^T \Bv - \rho^2) = 0
	\\
	&\lambda \geq 0
\end{align}

If the constraint is not active, then $\lambda = 0$ and
\begin{align*}
	\Bv_* = -\frac{1}{2} \Dv^{-1} \vr
\end{align*}

In practice, unless $\rho$ is large the constraint will always be active, in which case \ref{eq:kkt_diagonal} becomes

\begin{align*}
	&-4\vr-8 \Dv \Bv+2\lambda(\Bv) = 0
	\\
	&(\Bv^T \Bv - \rho^2) = 0
	\\
	&\lambda \geq 0
\end{align*}

this then becomes 

\begin{align*}
	&\Bv_* = 2\left(\lambda I - 4\Dv\right)^{-1} \vr
	\\
	&4\sum_i^d  \frac{(\vr^i)^2}{\left(\lambda - 4(\bv_0^i(\bv_0^i+\vr^\prime)\right)^2}= \rho^2
	\\
	&\lambda \geq 0
\end{align*}

\subsection{Metric~(\ref{eq:gmix_diagonal})}

We follow the same approach as in the previous section. The main difference will be in the form of the geodesics: $\uv(t) \odot \vv(t) = \left(\uv_0 \odot\vv_0\right) \odot(1+2\Bv t)$, where $\Bv^i= \frac{\eta_{\uv}^i}{\uv^i}=\frac{\eta_{\vv}^i}{\vv^i}$, as in the previous section. This essentially treats the two-layer neural network as if it were a single layer, with predictor $\beta = \uv \odot \vv$, that it then perturbs linearly to determine sharpness. For $\langle \cdot,\cdot \rangle^{\text{mix}}$, and denoting by $\bm{D}_{\bv} = diag(\bv_0^i)$:
\begin{align}\label{eq:mix_sharpness}
S_{max} &= \max\limits_{||\eta||^{\text{mix}} \leq \rho} 4 \left[\Bv^T (\bv_0-\bv_{*})+\Bv^T \bm{D}_{\bv}^2 \Bv \right]
\end{align}
We also have that
\begin{align}
    (||\eta||^{\text{mix}})^2 &= \left[\ldots+(\vv^i)^2(\eta_{\uv}^i)^2+(\uv^i)^2(\eta_{\vv}^i)^2+\ldots \right]
    \\
    &=\left[\ldots+(\vv^i)^2(\uv^i)^2\left(\frac{(\eta_{\uv}^i)^2}{(\uv^i)^2}+\frac{(\eta_{\vv}^i)^2}{(\vv^i)^2}\right)+\ldots \right]
    \\
    &= \left[\ldots+2(\vv^i)^2(\uv^i)^2 (\Bv^i)^2+\ldots \right] = ||2\bm{D}_{\bv_0} \Bv||_2
\end{align}

Substituting $2\bm{D}_{\bv_0} \Bv = \bm{\delta}$, Eq.~\ref{eq:mix_sharpness} becomes
\begin{align}
S_{max} &= \max\limits_{||\bm{\delta}|| \leq \rho} \left[\bm{\delta}^T (\bv_0-\bv_{*})+\bm{\delta}^T \bm{\delta} \right]
\end{align}

with the solution (up to constants)
\begin{align}
S_{max} = \rho||\bv_0-\bv_{*}||_2
\end{align}
\section{Geodesic Sharpness: $GL_1$ symmetry and Adaptive Sharpness}\label{ap:ignoring_corrections}

What happens if instead of a general $\text{GL}_{n}$ symmetry, we factor out a $\text{GL}_1$ re-scaling symmetry? That is, we identify, element-wise, $(\bar{x}, \bar{y}) \sim (\bar{x}^{\prime} \bar{y}^{\prime})$ if $\exists \alpha \in \sR_{*} = \sR \setminus \{0\}$ s.t. $\bar{x} = \alpha \bar{x}^{\prime}$ and $\bar{y} = \alpha^{-1} \bar{y}$.

This is the symmetry present in diagonal networks, and so we utilize the metric given by Eq.~\ref{eq:ginv_diagonal}, reproduced below for convenience of the reader:
\begin{align}
g\left[\left(\eta_{\uv}, \eta_{\vv}\right),\left(\nu_{\uv}, \nu_{\vv}\right)\right]
= \sum_{i=1}^d \frac{\eta_{\uv}^i \nu_{\uv}^i}{(\uv^i)^2} +\frac{\eta_{\vv}^i \nu_{\vv}^i}{(\vv^i)^2}
\end{align}
Note that this metric is equivalent to the following metric:
\begin{align}
g\left[\left(\eta_{\uv}, \eta_{\vv}\right),\left(\nu_{\uv}, \nu_{\vv}\right)\right]
=  g\left[\left(\eta_{\uv}/|\uv| , \eta_{\vv}/|\vv|\right),\left(\nu_{\uv}/|\uv|, \nu_{\vv}/|\vv|\right)\right]_{\text{euc}}
\end{align}
where $g_{\text{euc}}$ is the usual Euclidean metric and the division is taken to be element-wise. Denoting the concatenation of all tangent vectors by $\xi$, the concatenation of all parameters by $\vw$, we have $||\xi|| = ||\xi/|\vw|||_2$.

In this situation Eq.~\ref{eq:max-sharpness} becomes ($\gamma$ denotes our geodesics as usual)
\begin{align}
\begin{split}
	S_{\text{max}}^{\rho}(w,c) &= \mathbb{E}_{\sS \sim \sD} \left[ \max_{||\xi/|\vw|||_2 \leq \rho} L_{S}(\bar{\gamma}_{\bar{\xi}}(1)) - L_{S}(\bar{\gamma}_{\bar{\xi}}(0)) \right], 
 \end{split}
\end{align}
If we then ignore the corrections induced by the geometry of the metric on the geodesics, i.e., take $\bar{\gamma}_{\bar{\xi}}(1)=\bar{\gamma}_{\bar{\xi}}(0)+\bar{\xi}=\vw+\bar{\xi}$, then we get

\begin{align} \label{eq:max_def}
    S_{\text{max}}^{\rho}(w,c) &= \mathbb{E}_{\sS \sim \sD} \left[ \max_{||\xi/|\vw|||_2 \leq \rho} L_{S}(\vw+\xi) - L_{S}(\vw) \right]
\end{align}

which is exactly the formula for adaptive sharpness.
\section{Geodesic Sharpness: Ablations}\label{ap:ablations}
In this appendix we conduct ablation studies on geodesic sharpness (\cref{eq:max-sharpness}). There are two main components to our recipe that differ from adaptive sharpness: a) the norm $||\bar{\xi}||$; b) the weight update formula, which instead of the usual $\vw^i = \vw^i+\bar{\xi}$ takes into account the curvature induced by the parameter space symmetries $\vw^i = \vw^i+\bar{\xi}^i -\frac{1}{2} \Gamma^i_{kl}\bar{\xi}^k\bar{\xi}^l$.
Below we turn off these components one by one and re-compute the resulting sharpness on MNLI using the BERT models described in Section~\ref{sec:bert}.

\begin{figure}[!h]
\centering
\begin{minipage}{0.35\textwidth}
        \includegraphics[width=\textwidth]{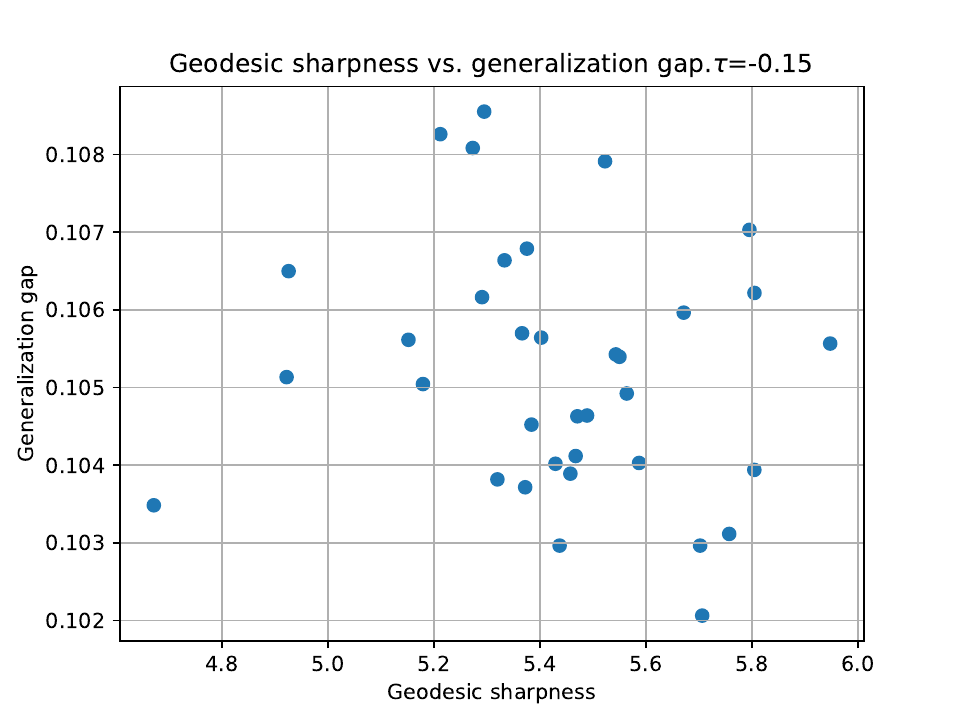}
\end{minipage}%
\begin{minipage}{0.35\textwidth}
        \includegraphics[width=\textwidth]{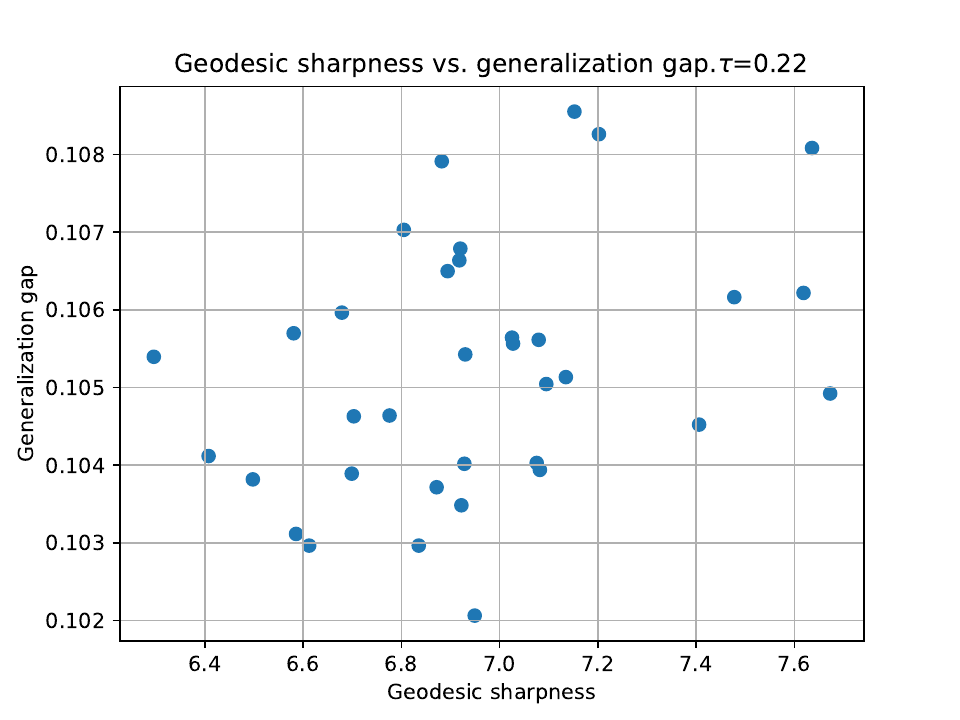}
\end{minipage}%
\begin{minipage}{0.35\textwidth}
        \includegraphics[width=\textwidth]{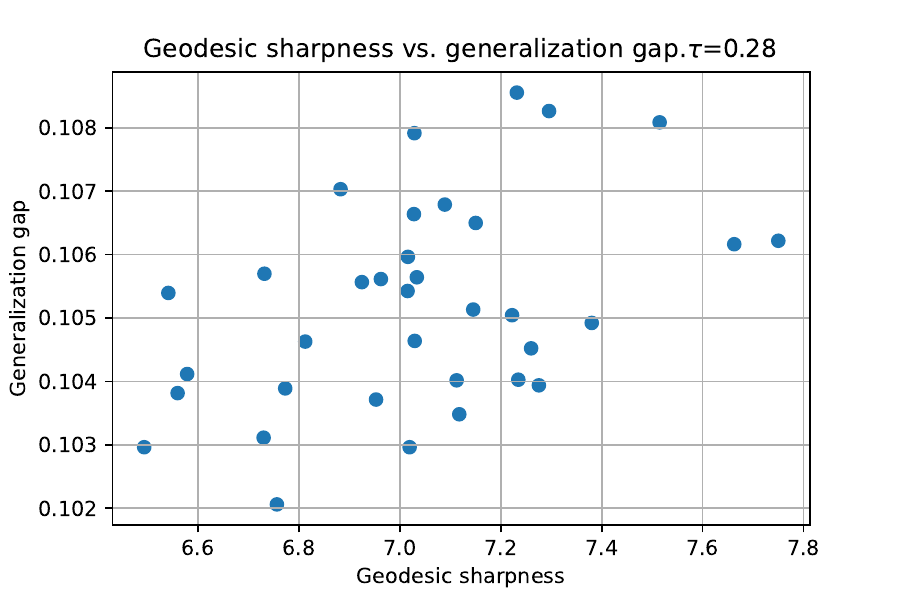}
\end{minipage}%
\caption{The generalization gap on the MNLI dev matched set \citep{mnli2018} vs. worst-case  adaptive sharpness with metric (\ref{eq:attention_metric}) is shown for 35 models from~\citep{mccoy2020bertsfeather}. On the left we plot the results when we turn off the corrected norm, and on the middle when we turn off the second-order weight corrections. Right are the results with no ablations.}\label{fig:metric_ablation}
\end{figure}

\paragraph{Metric (\ref{eq:attention_metric})}
In Figure~\ref{fig:metric_ablation} we show the results for our ablation studies using metric (\ref{eq:attention_metric}). The norm component is much more impactful than the second-order weight corrections. Turning off the second-order weight corrections results in a small performance drop only.

\begin{figure}[!h]
\centering
\begin{minipage}{0.35\textwidth}
        \includegraphics[width=\textwidth]{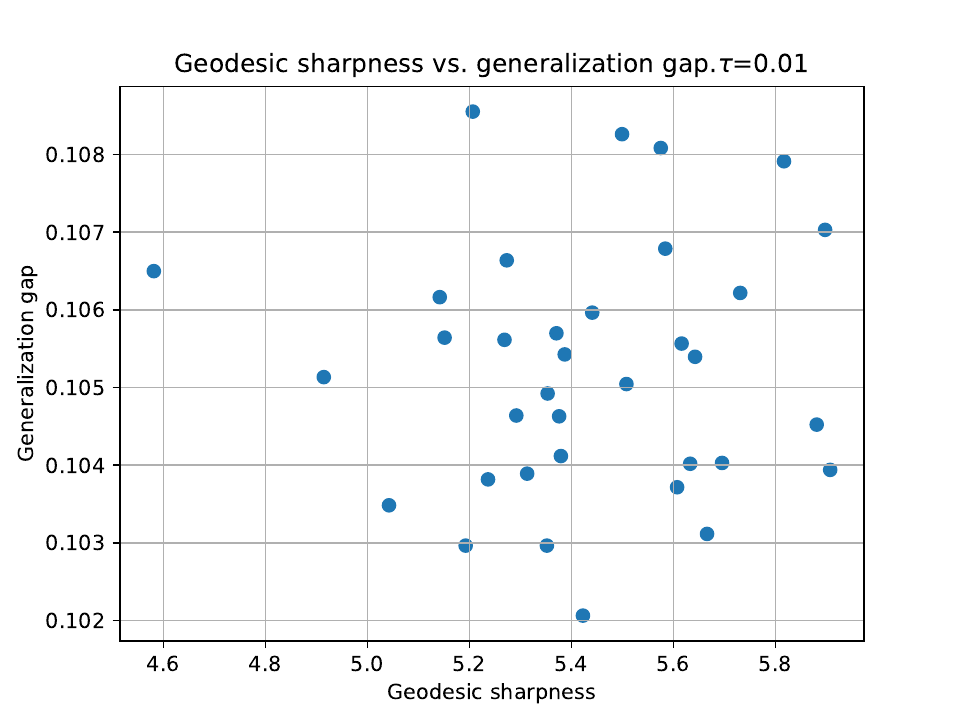}
\end{minipage}%
\begin{minipage}{0.35\textwidth}
        \includegraphics[width=\textwidth]{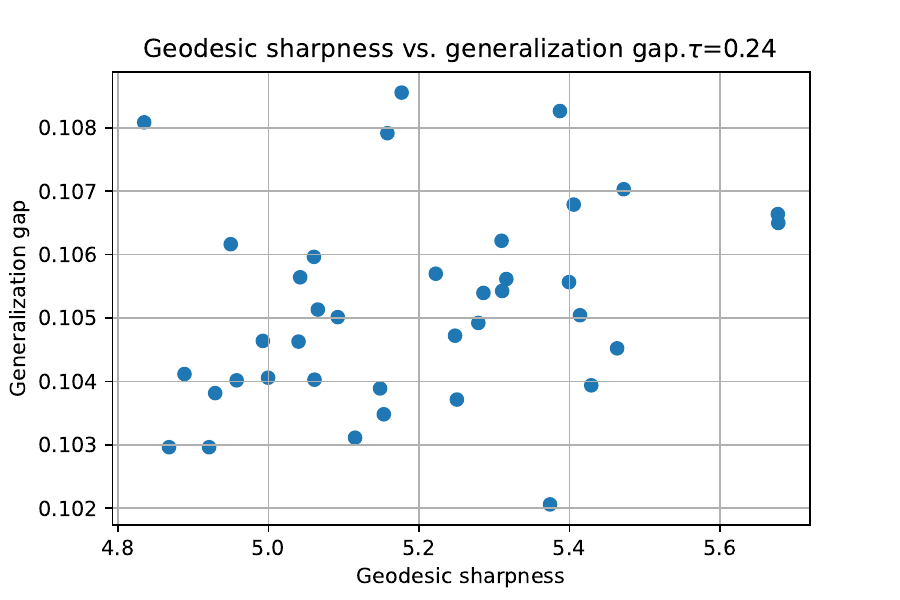}
\end{minipage}%
\begin{minipage}{0.35\textwidth}
        \includegraphics[width=\textwidth]{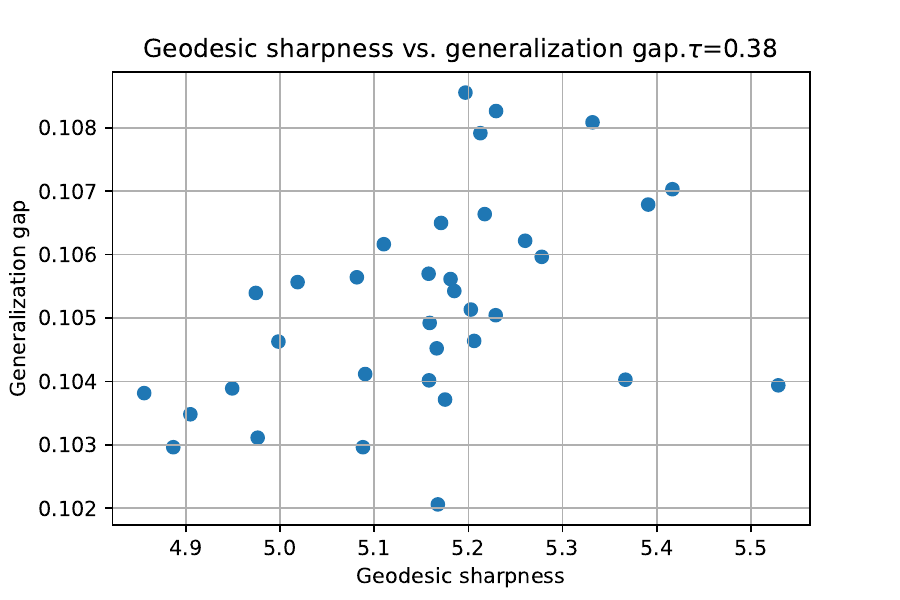}
\end{minipage}
\caption{The generalization gap on the MNLI dev matched set \citep{mnli2018} vs. worst-case  adaptive sharpness with metric (\ref{eq:attention_metric_mix}) is shown for 35 models from~\citep{mccoy2020bertsfeather}. On the left we plot the results when we turn off the corrected norm, and on the middle when we turn off the second-order weight corrections. On the right are the results with no ablations.}\label{fig:alternative_metric_ablation}
\end{figure}

\paragraph{Metric (\ref{eq:attention_metric_mix})}
In Figure~\ref{fig:alternative_metric_ablation} we show the results for our ablation studies using metric (\ref{eq:attention_metric_mix}). The norm component is still much more impactful than the second-order weight corrections for this metric, but now the second-order weight corrections are essential, and without them sharpness loses a considerable amount of predictive power.


\section{Geodesic Sharpness: Ranks and Relaxation}
\subsection{Ranks: how natural is Assumption 5.1?}
In general, in non-linear networks there is a tendency towards low-rank representations, which might make Assumption 5.1 seem excessive and counter to realistic situations. However, while the learned $W_Q W_K^T$ tend to be low-rank, $W_Q$ and $W_K$ (on which Assumption 5.1 ought to apply) themselves are usually high/full (column) rank~\citep{yu2023lowrank}.

\subsection{Relaxation}\label{ap:relaxation}
\begin{figure}[!h]
\centering
\begin{minipage}{0.5\textwidth}
        \includegraphics[width=\textwidth]{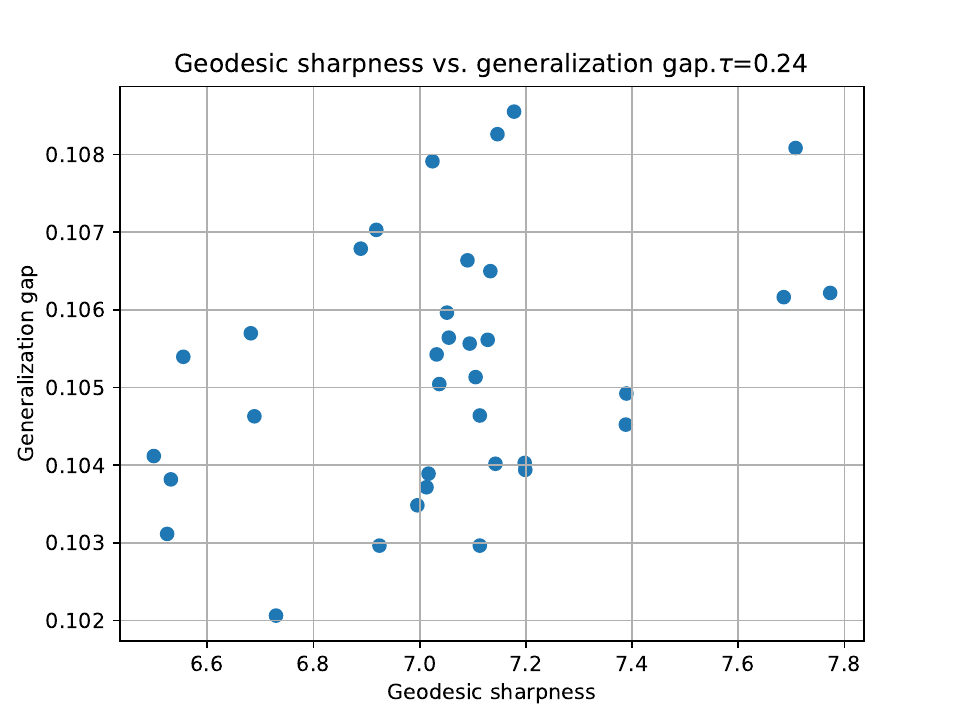}
        \caption*{Relaxation $=10^{-2}$}
\end{minipage}%
\begin{minipage}{0.5\textwidth}
        \includegraphics[width=\textwidth]{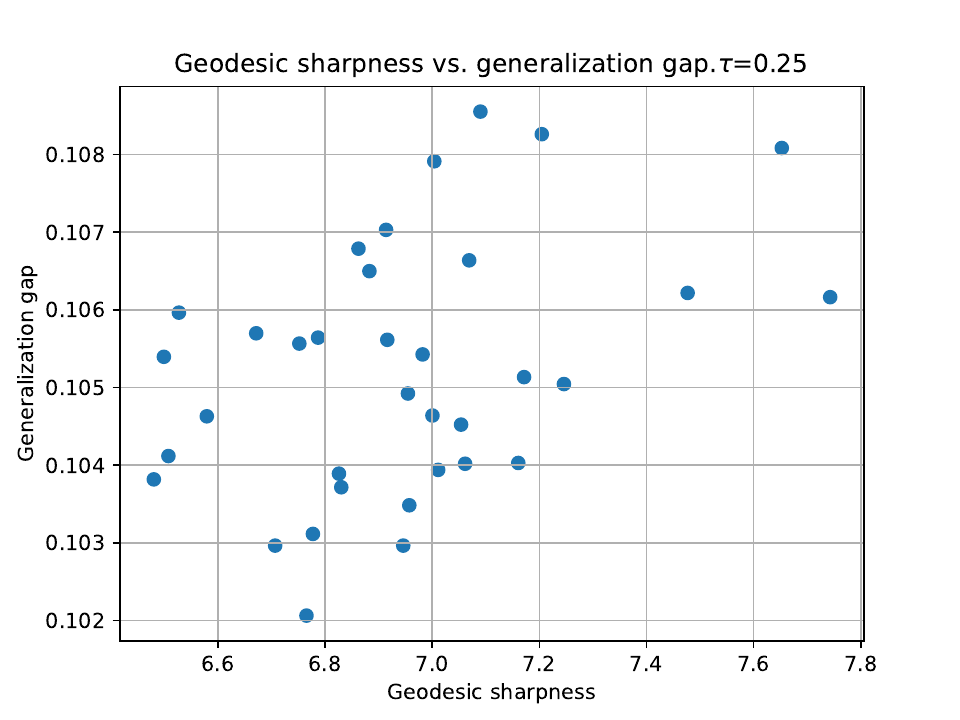}
        \caption*{Relaxation $=10^{-3}$}
\end{minipage}%
\\
\begin{minipage}{0.5\textwidth}
        \includegraphics[width=\textwidth]{fig/BERT/New/sinv.pdf}
        \caption*{Relaxation $=10^{-4}$}
\end{minipage}%
\begin{minipage}{0.5\textwidth}
        \includegraphics[width=\textwidth]{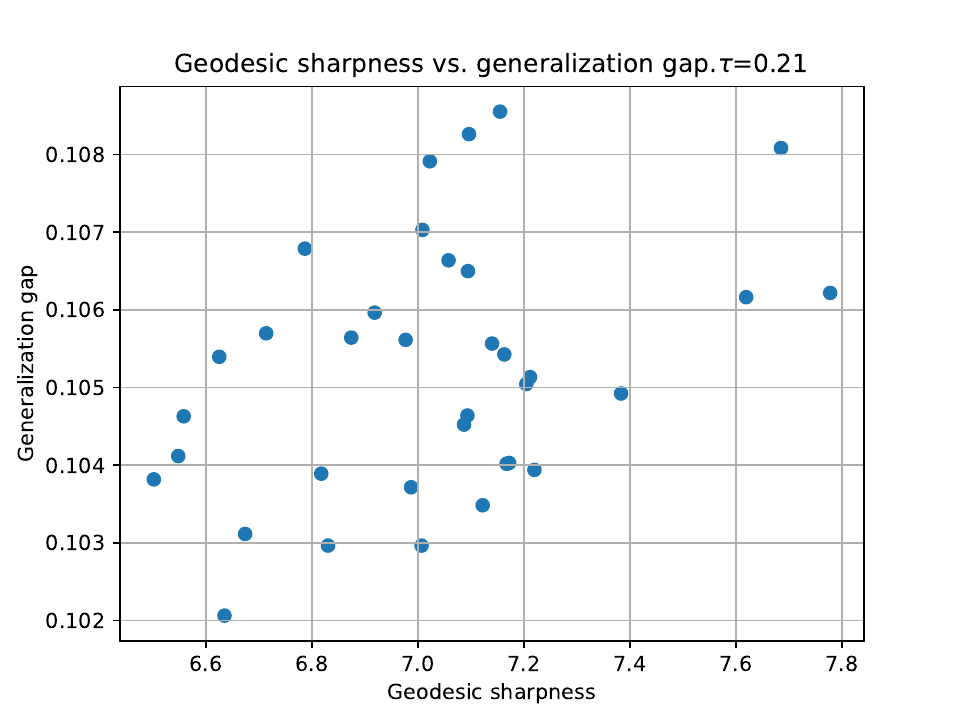}
       \caption*{Relaxation $=0$}
\end{minipage}
\caption{The generalization gap on the MNLI dev matched set \citep{mnli2018} vs. worst-case  adaptive sharpness (left) and geodesic sharpness ($\langle \cdot, \cdot \rangle^{\text{inv}}$), is shown for 35 models from~\citep{mccoy2020bertsfeather}. Only the relaxation parameter differs between plots. The results stay broadly the same.}\label{fig:relaxation}
\end{figure}

Due to the definition of metric~\ref{eq:attention_metric}, we need to invert matrices of the type of $W_Q^T W_Q$. When these are not full-rank, numerical stability can suffer. Due to floating-point precision rounding errors, in practice $W_Q^TW_Q$ is always invertible, but sometimes the inverted matrices have huge singular values. To combat this, we introduce a relaxation parameter, so that $W_Q^TW_Q\rightarrow W_Q^TW_Q+\epsilon I_h$, which dampens the resulting singular values. Although we cannot take it to be exactly zero, as long as it is small enough, numerical stability is improved and the results remain roughly the same. We study the effects of varying this parameter on our results empirically below (Figure~\ref{fig:relaxation}), using the same setup as in Section~\ref{sec:bert}. The results are not significantly affected by the variation of this parameter.

\section{Additional Derivations and Proofs}\label{ap:proofs}
\subsection{Proof that Eq.~\ref{eq:attention_metric} defines a valid Riemannian metric}\label{ap:metric_proof}
Eq.~\ref{eq:attention_metric} defines a valid metric on the total space $\overline{\gM}$ if it is smooth, and for each point $(\bar{G},\bar{H}) \in \overline{\gM}$ it defines a valid inner product on the tangent space $T_{(\bar{G},\bar{H})} \overline{\gM}$. 
That it is smooth is obvious, so we show that $\langle \bar{\eta}, \bar{\zeta} \rangle_{(\bar{G},\bar{H})}
    =
    \Tr
    \left(
    ( \mG^{\top}\mG )^{-1}
    \bar{\eta}_{\mG}^{\top} \bar{\zeta}_{\mG}
    +
    ( \mH^{\top}\mH )^{-1}
    \bar{\eta}_{\mH}^{\top} \bar{\zeta}_{\mH}
    \right)$ defines a valid inner product:
\begin{enumerate}[label=(\roman*)]
    \item {\it Symmetry} $\langle \bar{\eta}, \bar{\zeta} \rangle = \langle \bar{\zeta}, \bar{\eta} \rangle$:  omitting the $\mH$ term as it is identical, $\langle \bar{\eta}, \bar{\zeta} \rangle = \Tr
    \left( (\mG^{\top}\mG )^{-1}
    \bar{\eta}_{\mG}^{\top} \bar{\zeta}_{\mG}\right)=\Tr
    \left( (\mG^{\top}\mG )^{-1}
    \bar{\zeta}_{\mG}^{\top} \bar{\eta}_{\mG}\right)=\langle \bar{\zeta}, \bar{\eta} \rangle$ ;
    \item {\it Bilinearity} $\langle a\bar{\eta} + b\bar{\zeta}, \bar{\lambda} \rangle = a \langle \bar{\eta}, \bar{\lambda}  \rangle + b \langle \bar{\zeta}, \bar{\lambda} \rangle = \langle \bar{\lambda} , a\bar{\eta} + b\bar{\zeta} \rangle$: follows by linearity of the trace;
    \item {\it Positive Definiteness} $\langle \bar{\eta},\bar{\eta}\rangle \geq 0$: using assumption 5.1, $\mG^T \mG$ is invertible and is positive-definite; this means that $(\mG^T \mG)^{-1}$ is also positive-definite, and so $\langle \bar{\eta},\bar{\eta}\rangle \geq 0$, with equality only when $\bar{\eta} = 0$.
\end{enumerate}

The proof that~\cref{eq:attention_metric_mix} defines a valid metric is analogous.
\subsection{Derivation of the geodesic corrections for attention}\label{ap:proofs_attention_geodesic}

We apply the Euler-Lagrange formalism to the energy functional to derive the geodesic equation on the attention quotient manifold, and hence $\Gamma^i_{kl} \bar{\xi}^k_\mG \bar{\xi}^l_\mG$, remembering that geodesics, in local coordinates, obey the equation $\frac{d^2 \gamma^{i}}{dt^2} + \Gamma^i_{k l} \frac{d\gamma^k}{dt} \frac{d\gamma^l}{dt} = 0$.

Starting from 

\begin{align}\label{eq:energy_functional}
E(\gamma) &= \int_0^1 \mathcal{L}~dt = \int_0^1 \langle \dot{\gamma}(t),\dot{\gamma} (t)\rangle_{\gamma(t)} dt
\\ &= \int_0^1 \left[\Tr(\gamma_\mG(t)^T\gamma_\mG(t)) \dot{\gamma}_\mG(t)^T \dot{\gamma}_\mG(t)+\Tr(\gamma_\mH(t)^T\gamma_\mH(t)) \dot{\gamma}_\mH(t)^T \dot{\gamma}_\mH(t)\right] dt
\end{align},

The Euler-Lagrange equation, for the $\mG$ part only, reads

\begin{align} \label{eq:euler-lagrange}
    \frac{d}{dt}\left(\frac{\partial \mathcal{L}}{\partial \dot{\mG}}\right)-\frac{\partial \mathcal{L}}{\partial \mG} = 0
\end{align}

We have

\begin{align}
    &\frac{\partial \mathcal{L}}{\partial \mG} = -2 \mG \left(\mG^T \mG\right)^{-1}\left(\dot{\mG}^T \dot{\mG}\right) \left(\mG^T \mG\right)^{-1}\\
    &\frac{d}{dt}\left(\frac{\partial \mathcal{L}}{\partial \dot{\mG}}\right) = 2 \ddot{\mG} \left(\mG^T \mG\right)^{-1}-2 \dot{\mG} \left(\mG^T \mG\right)^{-1}\left(\dot{\mG}^T \mG+\mG^T \dot{\mG}\right)\left(\mG^T \mG\right)^{-1}
\end{align}

So that Eq.~\ref{eq:euler-lagrange} becomes:

\begin{align}
     \ddot{\mG}- \dot{\mG} \left(\mG^T \mG\right)^{-1}\left(\dot{\mG}^T \mG+\mG^T \dot{\mG}\right)+\mG \left(\mG^T \mG\right)^{-1}\left(\dot{\mG}^T \dot{\mG}\right) = 0
\end{align}

From which we read
\begin{align}
    \Gamma^i_{kl} \bar{\xi}^k_\mG \bar{\xi}^l_\mG = \left[- \bar{\xi} \left(\mG^T \mG\right)^{-1}\left(\bar{\xi}^T \mG+\mG^T \bar{\xi}\right)+\mG \left(\mG^T \mG\right)^{-1}\left(\bar{\xi}^T \bar{\xi}\right)\right]^i
\end{align}

The same reasoning is used to deduce Eq.~\ref{eq:attention_approximation_mix}.
\subsection{Metrics related by scaling and constants}\label{ap:metric_scale}
If $g$ is a metric and $g_{\text{scaled}}=Cg+D$, then from Eq.~\ref{eq:energy_functional} and Eq.~\ref{eq:euler-lagrange} we see that the geodesics induced by $g_{\text{scaled}}$ are the same as those induced by $g$. 
The geodesic sharpness induced by $g_{\text{scaled}}$ is 
\begin{align*}
\begin{split}
	S_{\text{max}}^{\rho}(w) &= \mathbb{E}_{\sS \sim \sD} \left[ \max_{|| \bar{\xi}||_{\bar{\gamma}_{\text{scaled}}} \leq \rho} L_{S}(\bar{\gamma}_{\bar{\xi};\text{scaled}}(1)) - L_{S}(\bar{\gamma}_{\bar{\xi};\text{scaled}}(0)) \right] =
    \\ &=\mathbb{E}_{\sS \sim \sD} \left[ \max_{C|| \bar{\xi}||_{\bar{\gamma}}+D \leq \rho} L_{S}(\bar{\gamma}_{\bar{\xi}}(1)) - L_{S}(\bar{\gamma}_{\bar{\xi}}(0)) \right],
    \\ &=\mathbb{E}_{\sS \sim \sD} \left[ \max_{|| \bar{\xi}||_{\bar{\gamma}} \leq \rho^{\prime}} L_{S}(\bar{\gamma}_{\bar{\xi}}(1)) - L_{S}(\bar{\gamma}_{\bar{\xi}}(0)) \right],
 \end{split}
\end{align*}
So they are the same up to some re-definition of the hyperparameter $\rho$.


\end{document}